\documentclass[10pt,twocolumn,letterpaper]{article}

\usepackage[pagenumbers]{cvpr} 

\usepackage{colortbl}

\definecolor{cvprblue}{rgb}{0.21,0.49,0.74}
\usepackage[pagebackref,breaklinks,colorlinks,allcolors=cvprblue]{hyperref}

\usepackage{xcolor}
\usepackage{footmisc}

\definecolor{customred}{HTML}{ED008C}

\def\paperID{3110} 
\def\confName{CVPR}
\def\confYear{2025}

\title{Graph Domain Adaptation with Dual-branch Encoder and Two-level Alignment for Whole Slide Image-based Survival Prediction}

\author{Yuntao Shou$^{1}$, PeiqiangYan$^1$, Xingjian Yuan$^1$, Xiangyong Cao$^{1,\dagger}$, Qian Zhao$^{1, \dagger}$, Deyu Meng$^1$ \\
$^1$Xi’an Jiaotong University, Xi’an, 710049, China \\ 
$\dagger$ Corresponding author \\
{\tt\small shouyuntao@stu.xjtu.edu.cn, Yim310@stu.xjtu.edu.cn, 2276215731@stu.xjtu.edu.cn,}\\
{\tt\small caoxiangyong@mail.xjtu.edu.cn, timmy.zhaoqian@xjtu.edu.cn, dymeng@mail.xjtu.edu.cn}
}

\newcommand{\zq}[1]{\textcolor{black}{#1}}

\begin{document}
\maketitle

\begin{abstract}
In recent years, \zq{histopathological whole slide image (WSI)-based survival analysis has attracted much attention in medical image analysis. In practice,} WSIs usually come from different hospitals or laboratories, \zq{which can be seen as different domains, and thus} may have significant differences in imaging equipment, processing procedures, and sample sources. \zq{These differences generally result in large gaps in distribution between different WSI domains, and thus the survival analysis models trained on one domain may fail to transfer to another.} 
To address \zq{this issue}, we propose a Dual-branch Encoder and Two-level Alignment (DETA) framework to explore \zq{both feature and category-level} alignment between different \zq{WSI} domains. Specifically, \zq{we first formulate the concerned problem as graph domain adaptation (GDA) by virtue the graph representation of WSIs. Then we construct a dual-branch graph encoder, including the message passing branch and the shortest path branch, to explicitly and implicitly extract semantic information from the graph-represented WSIs. To realize GDA, we propose a two-level alignment approach: at the category level, we develop a coupling technique by virtue of the dual-branch structure, leading to reduced divergence between the category distributions of the two domains; at the feature level, we introduce an adversarial perturbation strategy to better augment source domain feature, resulting in improved alignment in feature distribution.} 
\zq{To the best of our knowledge, our work is the first attempt to alleviate the domain shift issue for WSI data analysis. Extensive experiments on four TCGA datasets have validated the effectiveness of our proposed DETA framework and demonstrated its superior performance in WSI-based survival analysis.}
\end{abstract}

\begin{figure}[t]
    \centering
    \includegraphics[width=1\linewidth]{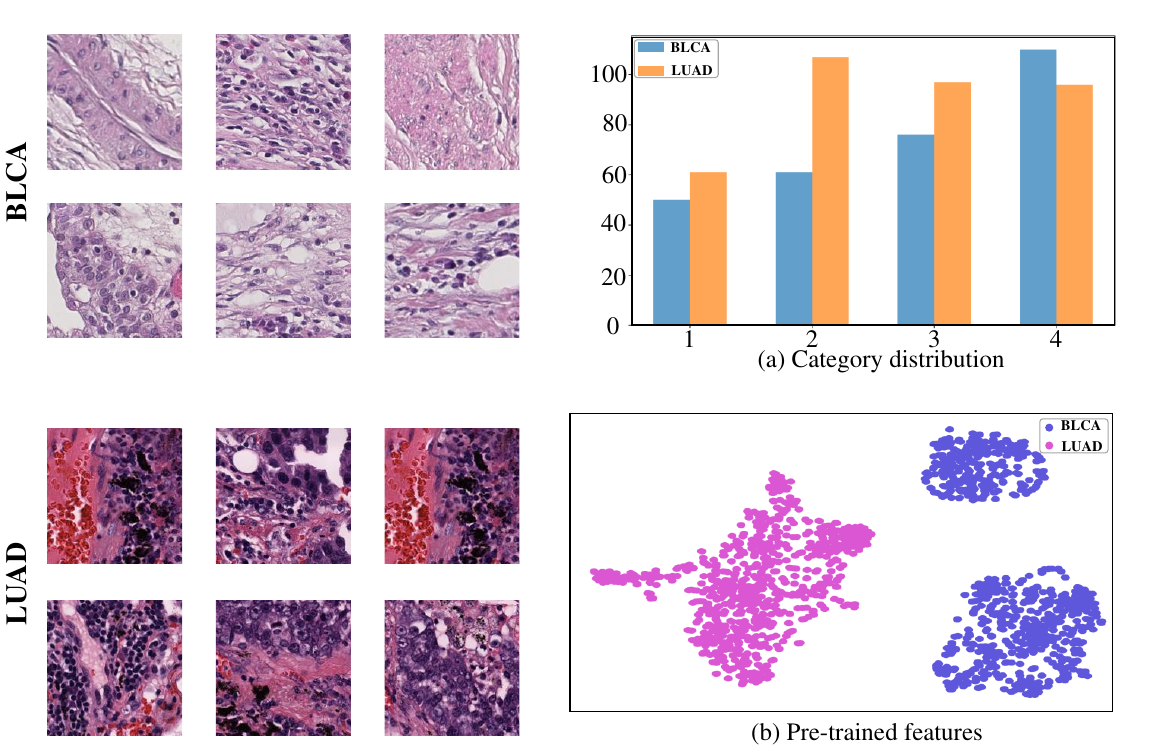}
    \vspace{-7mm}
    \caption{\zq{An example of pathology images with distribution shift. The left shows image from two WSI datasets, i.e., BLCA and LUAD. The right shows the category distribution and the t-SNE visualization of the extracted features by CLAM \cite{lu2021data} method.}}
    \label{fig:enter-label}
    \vspace{-5mm}
\end{figure}

\section{Introduction}
\zq{In clinic, survival analysis aims to predict the time from the start of treatment to death or disease recurrence, which plays an important role in clinical prognosis \cite{jaume2024modeling}. Specifically, accurate survival prediction not only helps doctors understand the progression of the disease but also evaluates the effectiveness of different treatments, which essentially impact patients' clinical outcomes \cite{jaume2024transcriptomics}.} Traditional survival analysis methods usually rely on short-term clinical indicators \cite{kalafi2019machine, yu2021dynamic} and long-term follow-up reports \cite{capra2017assessing, vale2021long}, which are often very time-consuming in practical applications. In recent years, with the development of deep learning technology, significant progress has been made in medical image analysis, \zq{allowing survival analysis to be more effectively integrated with more informative pathology images \cite{sormani2015can, inusah2010assessing}. By virtue of deep neural networks (DNNs), features from pathological images can be automatically extracted, thereby helping researchers to establish powerful survival models. In recent years, the whole slide histopathological images (WSIs) are of particular interest for this task \cite{baroncini2019long, ruggieri2024assessing, yperman2022motor}, since they can provide rich information of diseased tissue samples, e.g., the microscopic changes in tumor cells and their microenvironment, allowing the model to have a deeper understanding of the characteristics and behavior of the tumor.}

\zq{Since WSIs are generally very large, e.g., can be with 1 million by 1 million pixels, a common strategy in WSI-based survival analysis is to divide a WSI into patches sharing a same label, which can thus be formulated as a multi-instance learning (MIL) problem \cite{ahmedpathalign, chen2024towards, srinidhi2021deep, nakhli2023co}. By exploiting the similarities between those patches, a graph with patches being its nodes can be constructed as a representation of the WSI, and thus the graph-based methods relying on the graph convolutional networks (GCNs) \cite{shou2022conversational, shou2025masked, shou2022object, shou2023comprehensive, shou2024adversarial, meng2024deep, shou2023adversarial, ai2023gcn} have been developed to solve such a MIL problem \cite{zhao2020predicting, li2018graph}. Since GCNs can effectively capture complex spatial structures and contextual information to reflect potential pathological features and their interactions \cite{ai2024gcn, meng2024multi, shou2024contrastive, shou2024spegcl, shou2024efficient, ying2021prediction}, a good global representation of the WSI is expected to be learned, thereby achieving accurate survival prediction among the state-of-the-art ones \cite{di2022generating, wang2024dual}.}

\zq{Though achieving promising performance, existing WSI-based survival analysis methods are generally constructed without considering the domain shift issue which is often encountered in real scenarios. Specifically, WSIs collected by different hospitals or laboratories may have significant differences in distribution due to imaging equipment, processing procedures, and sample sources. We show an example in Fig. \ref{fig:enter-label} by visualizing the category distributions and the distributions of features extracted by CLAM \cite{lu2021data} of two WSI datasets. It can be seen that, in addition to the apparent visual differences in the original patches shown on the left, both the category and feature distributions of the two datasets significantly deviate from each other. Due to such a domain shift, the survival analysis models trained on one domain could hardly transfer to another, resulting in degraded performance in the target domain, which limits their availability in real applications. 
}

\zq{To address the above issue, we propose an innovative Dual-branch Encoder and Two-level Alignment (DETA) framework to realize the graph domain adaptation (GDA) for the WSI-based survival analysis, by fully exploring both category and feature-level distribution alignment between different WSI domains. In specific, using the graph representation of WSIs, we first formulate the concerned problem as the GDA. Then we design a dual-branch graph encoder, containing a message passing (MP) and a shortest path (SP) branch to fully utilize the features of both the source and target WSI graphs. Within the dual structure, the MP branch can implicitly learn the topological semantics in the graph by neighborhood aggregation, while the SP branch complementarily provides explicit high-order structural semantics. Next, we develop several techniques to align the source and target domains at both category and feature levels. Specifically, at the category level, we define coupled evidence lower bounds, by virtue of the dual branch structure of the graph encoder to jointly train the graph encoder and risk predictor, so that the divergence between the category distributions of the two domains is expected to be reduced; at the feature level, we adversarially train a domain classifier and learn the perturbations to the graph features of the source domain, so that the feature distributions of both domains can be better aligned. With all these designs, the final survival prediction performance is expected to be enhanced.}

The main contributions of this paper can be summarized as follows:
\zq{
\begin{itemize}
    \item To the best of our knowledge, we make the first attempt to alleviate the domain shift issue in the WSI-based survival analysis task, thereby enhancing its availability in real scenarios.
    \item We propose a Dual-branch Encoder and Two-level Alignment (DETA) framework to explicitly and implicitly extract semantic information from WSIs, as well as align both the category and feature distributions of source and target domains, such that the risk predictor can perform well in the target domain. 
    \item We apply the proposed to four public TCGA datasets with different adaptation directions, and demonstrate its superior performance for the WSI-based survival analysis.
\end{itemize}
}

\section{Related Work}

\subsection{WSIs-based Survival Analysis}
Survival prediction plays a vital role in clinical medicine, providing doctors with valuable information to evaluate disease progression and treatment effects. Traditionally, survival prediction usually relies on rich clinical data, including short-term clinical indicators \cite{kalafi2019machine}, \cite{yu2021dynamic}, long-term follow-up reports \cite{capra2017assessing, vale2021long} and various radiological images \cite{francone2020chest, platz2017dynamic}. \zq{Recently, with the rapid development of deep learning, WSI-based survival analysis has attracted increasing attention, since rich information about the disease can be provided by the gigapixel pathological images \cite{chen2022scaling, lu2023visual}. Since the WSI is generally with very high resolution, a common strategy is to divide it into patches sharing the same labels and treat the WSI-based survival analysis as MIL problem \cite{ahmedpathalign, zhao2023survival}.} For example, CMIB \cite{zhouyou2024cmib} captures the private and public features of different modalities through a co-attention mechanism and introduces a multimodal information bottleneck to obtain a robust representation of fused features. TANGLE \cite{jaume2024transcriptomics} adopts a modality-specific encoder and contrastive learning to aligns the outputs and learn a task-agnostic slice embeddings. \zq{However, since WSIs come from different hospitals or laboratories, the domain shift issue can occur in real scenarios, limiting the availability of WSI-based survival analysis in practical applications, which has not yet been addressed in current studies.}


\subsection{Unsupervised Domain Adaptation}
Unsupervised domain adaptation (UDA) aims to learn domain-invariant representations so that the model can be transferred from a source domain with rich labels to a target domain with scarce labels, which is crucial to improving the generalization ability of the model in different application scenarios \cite{ma2021active}, \cite{singh2021clda}, \cite{fengtowards}, \cite{mancini2019inferring}. Existing UDA methods can be roughly divided into two categories: domain difference-based methods and adversarial methods. Domain difference-based methods usually combine multiple distribution indicators (e.g., maximum mean difference \cite{saito2018maximum}  and Wasserstein distance \cite{shen2018wasserstein}) to measure the difference between the source domain and the target domain. By quantifying the distribution difference between the source domain and the target domain, the model is adjusted to better adapt to the target domain. In contrast, adversarial methods \cite{yin2023coco, dai2022graph} adopt a more implicit strategy to reduce inter-domain differences by introducing domain discriminators.


\subsection{Graph Domain Adaption}
Graph domain adaptation (GDA) focuses on how to effectively transfer information from source graphs to unlabeled target graphs to learn effective node-level representations \cite{lin2023multi}, \cite{wuhandling}, \cite{luo2023source}. Existing methods usually focus on domain alignment through graph neural networks (GNNs) \cite{shou2023graph, shou2023czl, meng2024masked, shou2024revisiting,ai2024edge,zhang2024multi, ai2023two} to promote knowledge transfer between source and target graphs \cite{wu2022attraction}, \cite{zhu2021transfer}, \cite{dai2022graph}. However, these methods often ignore the importance of category distribution alignment in the case of label scarcity and domain shift \cite{guo2022learning}, \cite{yin2023coco}, \cite{yehudai2021local}, \cite{yang2020heterogeneous}. Particularly, when the number of samples in the target domain is scarce, it is difficult for a model to accurately capture the feature distribution without sufficient category information.

\begin{figure*}
	\centering
	\includegraphics[width=1\linewidth]{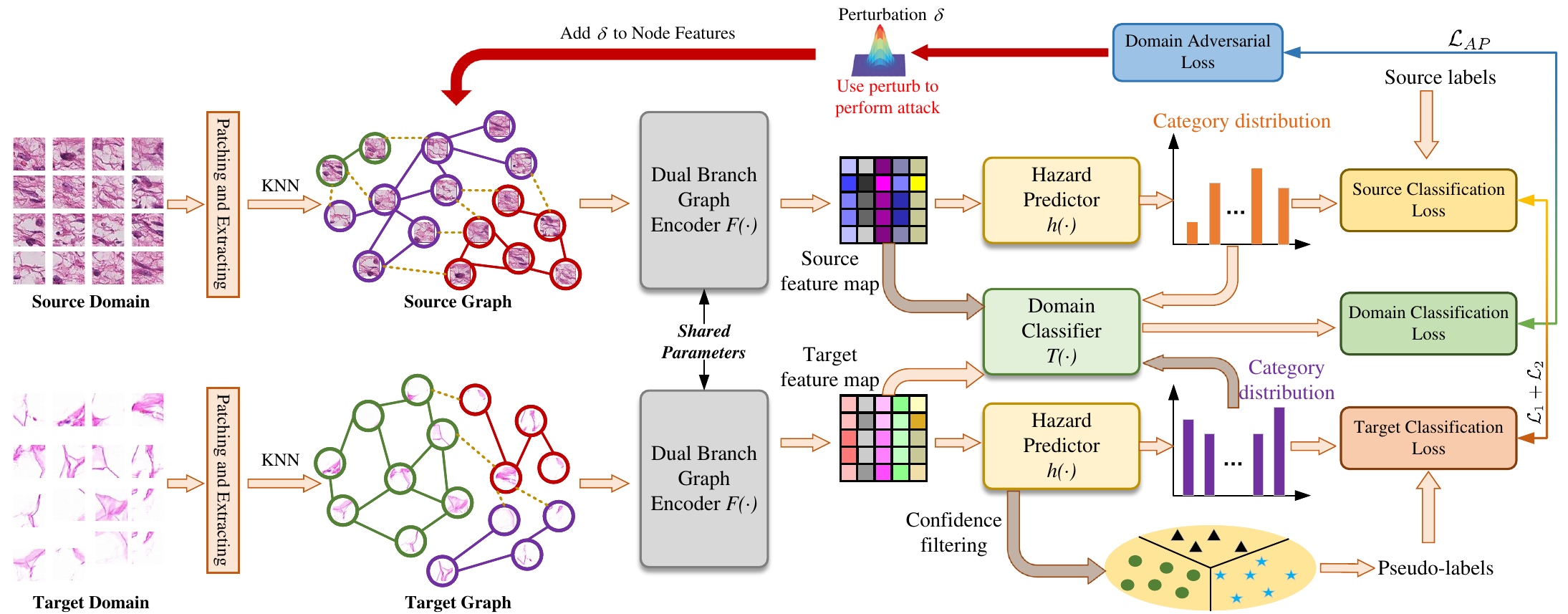}
	\caption{An overview of our proposed archicture DETA. The dual
		branch graph encoder utilizes the MP branch and the SP branch to explicitly and implicitly extract semantic information. To realize GDA, we propose a two-level alignment strategy to train the graph encoder and the hazard predictor, using both the labeled WSI graphs from the source domain and the unlabeled ones from the target domain.}
	\label{fig:archi}
        \vspace{-4mm}
\end{figure*}

\section{\zq{Problem Formulation}}

\zq{\textbf{WSI-based survival analysis.} Suppose $Z=\{z_1, z_2, \ldots, z_N\}$ is the WSI-based survival data of $N$ patients, where $z_i$ is represented by a triplet $z_i=\{x_i, c_i, y_i\}$, $x_i$ is the WSI, $c_i\in \{0, 1\}$ denotes the right censorship status, with $c_i=1$ indicating that the $i$-th sample is uncensored and $c_i=0$ censored, and $y_i$ represents the observed time. The goal of survival analysis is to estimate the following hazard function $h(y|x)$:
\begin{equation}
\small
	\begin{aligned}h(y|x)&=\lim_{\Delta y\to0}\frac{P(y\leq O\leq y+\Delta y|O\geq y,x)}{\Delta y},
\end{aligned}\end{equation}
where $O$ is the survival time, and consequently obtain the survival function $S(y|x)=P(O\geq y|x)$.}

\zq{In the deep learning practice for the WSI-based survival analysis, the time is often discretized into intervals and $y_i$ takes the integer value indicating which time interval (or the risk level) the $i$-th patient belongs to \cite{zhao2023survival, zhou2023cross}. In this case, the hazard function becomes
\begin{equation}
\small
    h(y|x)=P(O=y|O\geq y,x),~~~~y=1,2,\cdots,K,
\end{equation}
where $K$ denotes the total risk levels, and then the survival function can be obtained as
\begin{equation}
\small
    S(y|x)=\prod\nolimits_{j=1}^y(1-h(j|x)).
\end{equation}
By virtue of DNNs, one can explicitly parameterize $h(y|x)$:
\begin{equation}
\small
    h(y|x)=\phi_y(g(x)),~~~~y=1,2,\cdots,K,
\end{equation}
where $g(\cdot)$ is the feature extractor and $\phi_y(\cdot)$ maps the feature vector to the conditional probability $P(O=y|O\geq y,x)$ so that $\sum_{y=1}^KP(O=y|O\geq y,x)=1$, which can be realized by a multilayer perception (MLP) with the softmax output. Then $h$ can be learned by minimizing the following loss derived from the maximum likelihood estimation:
\begin{equation}
\small
    \begin{split}
        \mathcal{L}_{surv}=&-\sum\nolimits_{i=1}^Nc_i\left[\log S(y_i|x_i)+\log h(y_i|x_i)\right]\\
                           &-\sum\nolimits_{i=1}^N(1-c_i)\log S(y_i+1|x_i)    
    \end{split}.\label{eq:l_sur}
\end{equation}
It should be mentioned that though many previous studies considered continuous time, we follow the discrete-time setting, since the category distribution can be easier to explore for domain adaptation.
}




\noindent\zq{\textbf{Graph representation for WSI-based survival analysis.} As mentioned in the Introduction, graph-based methods \cite{meng2024revisiting, ai2024mcsff, shou2023graphunet, ai2024graph, shou2024low} have achieved promising performance for WSI-based survival analysis \cite{zhao2020predicting, li2018graph, di2022generating, wang2024dual}, since the graph-representation is helpful to capture complex structures and contextual information to reflect pathological features and their interaction, leading to good global representations for WSIs. The main idea is to first divide the large WSI into a series non-overlap patches and extract features for them, and then construct a graph with each node being the feature of one patch and each edge represents the neighborhood relationship between two patches. Such a WSI graph can be formally represented as $G = (V, E, F)$, where $V$ denotes the node set, $E$ represents the edge set, and $F \in \mathbb{R} ^{|V| \times d}$ is the node attribute matrix, with $|V|$ is the total number of nodes, and $d$ is the dimension of the attribute (or feature) of each node. With the constructed WSI graphs, the original WSI-based survival analysis now can be regarded as a graph learning problem using data triplets $\{G_i,c_i,y_i\}_{i=1}^N$, where $G_i$ is the graph constructed by WSI $x_i$. In this work, following \cite{xiong2024mome, jaume2024modeling}, we adopt CLAM \cite{lu2021data} to crop each WSI into 512×512 non-overlap patches and extract 1024-dimensional feature for each patch, and use the K-nearest neighbor (KNN) algorithm to construct the graph according to the Euclidean distance between patch features.}



\noindent\textbf{GDA for survival analysis.} \zq{As mentioned in the Introduction, the above-formulated WSI-based survival analysis may suffer from the domain shift issue in real applications. Therefore, our work is to develop a GDA method to alleviate this issue. Before that, we first formally describe the problem we are trying to solve. Suppose we have a set of training data from the so-called source domain, with known labels of the risk level, denoted as $\mathcal{D}^s = \{G^s_i, c^s_i, y^s_i\}^{N_s}_{i=1}$, where $N_s$ is the number of data from the source domain. Then we can train a model to define the hazard function $h$ using $\mathcal{D}^s$. Now a new dataset comes from the so-called target domain with unknown labels, denoted as $\mathcal{D}^t = \{G^t_j\}^{N_t}_{j=1}$, where $N_t$ is the number of data from the target domain, and we want to predict the risk level for each $G^t_j$ using the trained $h$. However, since there is no guarantee that $\mathcal{D}^s$ and $\mathcal{D}^t$ have the same distribution, directly using $h$ could have poor prediction performance. Therefore, we try to fine-tune $h$ by aligning the distributions of the two domains to enable it to perform well on the target domain, which can be regarded as an unsupervised GDA task.}


\section{Proposed Method}
\zq{In this section, we first briefly overview the proposed DETA framework, and then describe the details of its components, as well as the training strategy.
\subsection{Fraemwork Overview}
Fig. \ref{fig:archi} shows the overview of the proposed DETA framework. In the specific, the main network consists of a dual-branch graph encoder $F(\cdot)$ that intends to both explicitly and implicitly extract semantic information, respectively using the SP branch and the MP branch, from the WSI graph, which will be discussed in detail in Section \ref{sec:graph_enc}; and a hazard predictor $h(\cdot)$ that takes the feature output by $F$ and predicts the risk level of the corresponding WSI, which is constructed with MLP. To realize GDA, we propose a two-level alignment strategy to train the graph encoder and the hazard predictor, using both the labeled WSI graphs from the source domain and the unlabeled ones from the target domain. Specifically, at the category level, we design the coupling dual evidence lower bounds (ELBOs), by virtue of the dual-branch structure of the graph encoder, to align the category distributions of the two domains; while at the feature level, we propose an adversarial perturbation strategy to add perturbations to the features extracted from the source domain in an adversarial way against that from the target one with the assist of the domain classifier $T(\cdot)$, such that the features distributions of the two domains are expected to be better aligned. The details of the category-level and feature-level alignment are to be discussed in Section \ref{sec:cat_align} and \ref{sec:feat_align} respectively.
}

\subsection{Design of the Dual-branch Graph Encoder}\label{sec:graph_enc}
\begin{figure}
    \centering
    \includegraphics[width=1\linewidth]{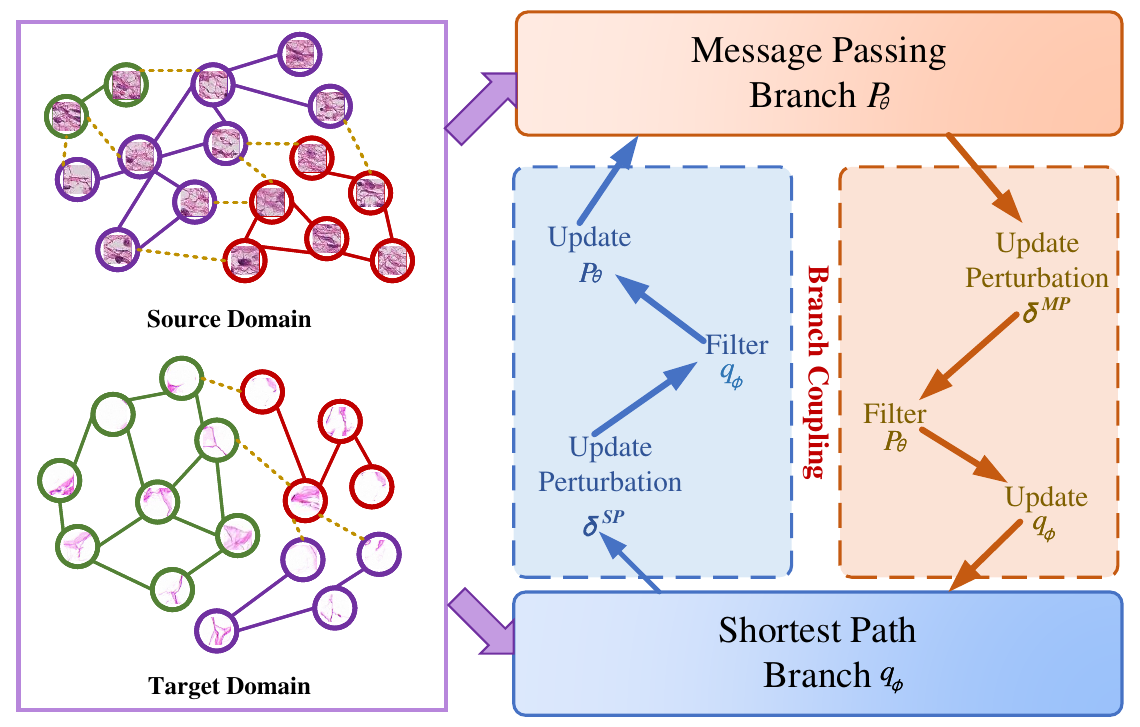}
    \caption{An overview of the proposed branch coupling. We adversarially optimize the perturbations to align domain distribution, and adopt an alternating strategy to align the category distribution.}
    \label{fig:dual}
    \vspace{-4mm}
\end{figure}

\zq{In existing studies, message passing neural networks (MPNNs) are often used to implicitly capture the topological semantics in the graph through neighborhood aggregation and use these representations to transfer knowledge between different domains \cite{yehudai2021local, sun2022gppt}, which have shown their effectiveness. However, since both the graph structures and feature distributions of different domains can largely deviate from each other, only relying on the neighborhood information could not be sufficient enough to capture the graph semantics for GDA. Therefore, we adopt the graph kernel method to explicitly capture the high-order semantic relation in the graph \cite{gao2021topology, borgwardt2005shortest} using the SP branch. Combining the MP branch, the dual-branch graph encoder is constructed, as shown in Fig. \ref{fig:dual}. In the following, we discuss the detail of the two branches.}

\noindent\textbf{The MP Branch.} \zq{The MP branch in our graph encoder consists of multiple stages of message passing process. One single message passing stage can be typically divided into two steps, i.e., message passing and node update. In the message passing step, the node $v$ receives messages, i.e., features, from its neighbor nodes $\mathcal{N}(v)$ and aggregates them; and in the node update step, the feature of each node is updated by combining its current feature and the aggregation of neighbor features. The whole process can be written as
\begin{equation}
\small
	h_v^{(t+1)}=\text{UPD}\left(h_v^{(t)},\text{AGG}\left(\left\{h_u^{(t)}:u\in\mathcal{N}(v)\right\}\right)\right)
\end{equation}
where $h_v^{(t)}$ denotes the feature of node $v$ at the $t$-th stage,} $\text{UPD}$ is the update function, and $\text{AGG}$ is an aggregation function, following the standard GCN process \cite{yin2023coco}.

\noindent\textbf{The SP Branch.} \zq{As mentioned before, the SP branch aims to explicitly capture the high-order semantic relation in the graph using the graph kernel method \cite{gao2021topology, borgwardt2005shortest}.}
Specifically, we generate a series of shortest paths from local substructures and extract higher-order and global semantic representations from the local information.

Let $\mathcal{N}_k(u)$ denote the set of all nodes reachable from node $u$ via the shortest path of length $k$. \zq{In each layer of the SP branch, the representation of node $u$ is updated by aggregating features all nodes from $\{\mathcal{N}_k(u)\}_{k=1}^K$,} where $K$ is a hyperparameter representing the maximum length of the shortest path. 
Furthermore, we introduce position encoding for the node representation ${m}_{u\in\mathcal{N}_k(u)}^l$ in each layer $l$ to identify the node's position as follows:
\begin{equation}
\small
	\begin{aligned}
	\hat{{m}}_{u\in\mathcal{N}_k(u)}^l&=\sigma({m}_{v\in\mathcal{N}_k(u)}^l+TE(k)), \\
	TE(k)_{2i}=\sin&\left(\frac{k}{10000^{2i/d}}\right),  
	TE(k)_{2i+1}=\cos\left(\frac{k}{10000^{2i/d}}\right),
	\end{aligned}
\end{equation}
where $\sigma$ is the activation function, $TE(k)$ is the position embeddings and $i \in [1, d]$ is the index of embeddings.

Then the update of the feature representation of node $v$ according to different path lengths $k$ can be summarized as
\begin{equation}
\small
	{m}_{u,\mathcal{N}_{k}(u)}^{l}=\mathcal{C}_{SP}^{l}\left({\hat{m}}_{u}^{l-1},\mathcal{A}_{SP}^{l}\left(\sum_{k=1}^{K}\left\{{\hat{m}}_{u}^{l-1}\right\}_{u\in\mathcal{N}_{k}(u)}\right)\right)
\end{equation}
where $\mathcal{C}_{SP}$ and $\mathcal{A}_{SP}$ is the combination and aggregation operators, respectively.

\subsection{\zq{Coupling Dual Branches for Category-level Alignment}}\label{sec:cat_align}
\zq{Generating pseudo labels for the target domain has been shown to be helpful for domain adaptation \cite{yehudai2021local, ding2021closer}. However, due to the domain shift issue, the pseudo labels could be inaccurate and have a category distribution different from that in the source domain. Therefore, by exploiting the characteristics of the dual-branch encoder structure, we proposed coupling the two branches by deriving ELBOs from two directions, to better align the category distributions of the two domains and enhance the quality of the pseudo label. In addition, we also use a filtering mechanism to retain high-confidence pseudo labels. It should be mentioned that, since the losses guide the training of the encoder, such a category-level alignment can also help align the features of two domains.}


\zq{Now we introduce the idea of our coupling strategy. Let's first consider the GDA problem we face: given the source graph $G^s$, its label $y^s$ and target graph $G^t$, we want to predict the label for $G^t$, denoted as $\hat{y}^t$, which can be treated as a latent variable here. Then we can write the following log-likelihood and its ELBO:
\begin{equation}
\small
	\begin{aligned}
		&\log p\left(y^s\mid G^s,G^t\right) \\ &\geq 
		\mathbb{E}_{q(\hat{y}^t\mid G^t)}\left[\log p\left(y^s,\hat{y}^t\mid G^s,G^t\right)
		-\log q\left(\hat{y}^t\mid G^t\right)\right],
	\end{aligned}
	\label{eq:9}  
\end{equation}
where the $p,q$ can be seen as two parameterizations for the label distribution given a graph $G$. Traditionally, such an ELBO is used to learn the approximation $q$ to the target $p$. However, note that in our model, there are two branches in the graph encoder, and thus we can use the two branches to parameterize the two distributions, respectively. In addition, we can also exchange the positions of $p$ and $q$, and thus the two branches can mutually act as targets and approximations to couple their training, thereby promoting the training efficacy.}

\zq{According to \cite{yin2024dream}, with $p$ being fixed, the ELBO in Eq. \eqref{eq:9} can be equivalently write as}
\begin{equation}
\small
	\begin{aligned}
		&KL\left(q\left(y^t\mid G^t\right)||p\left(y^t\mid G^s,G^t,y^s\right)\right)\\&=\mathbb{E}_{q(y^t\mid G^s,G^t,y^s)}[\log p\left(y^t\mid G^t\right)] + \mathbb{E}_{q(y^s,G^s)}\log q\left(y_{i}^{s}\mid G_{i}^{s}\right)
	\end{aligned}
    \label{eq:10}
\end{equation}
Then denoting the distributions parameterized by MP and SP as $p_{\theta}$ and $q_{\phi}$, using the coupling strategy introduced before, we get the following losses for updating MP and SP, respectively, with the other fixed:

\begin{equation}
\small
	\begin{aligned}
	\mathcal{L}_{1}=&-\mathbb{E}_{p_{\theta}\left(\hat{y}_{i}^{t}|G^{s},G^{t},y^{s}\right)>\zeta}\left[\log q_{\phi}\left(\hat{y}_{i}^{t}\mid G_{i}^{t}\right)\right] \\&-\mathbb{E}_{p_{\theta}(y^s,G^s)}\log p_{\theta}\left(y_{i}^{s}\mid G_{i}^{s}\right) \\
	\mathcal{L}_2=&-\mathbb{E}_{q_\phi\left(\hat{y}_i^t|G^s,G^t,y^s\right)>\zeta}\left[\log p_\theta\left(\hat{y}_i^t\mid G_i^t\right)\right]\\&-\mathbb{E}_{q_\phi(y^s,G^s)}\log q_\phi(y_i^s\mid G_i^s)
	\end{aligned}
\end{equation}
where $\hat{y}_{i}^{t}$ is the target graph pseudo-label filtered by the MP or SP branch. \zq{Note that in the above losses, we have introduced a threshold $\zeta$ when calculating the expectations, which acts as a confidence filter to select samples with high quality pseudo labels.}

\begin{table*}[htbp]
\footnotesize
	\centering
	\caption{The experimental results of survival analysis in one TCGA datasets as the training set and the other three datasets as the test sets. We highlight the top two best performing scores in red and blue, respectively.}
	\label{tab:BLCA}
    \renewcommand{\arraystretch}{0.9}
	\setlength{\tabcolsep}{2.2mm}{
		\begin{tabular}{lcccc|cccc}
			\toprule
                 \rowcolor{gray!20}
			Methods  &Year  & BLCA$\rightarrow$LGG & BLCA$\rightarrow$UCEC & BLCA$\rightarrow$LUAD & LGG$\rightarrow$BLCA & LGG$\rightarrow$UCEC & LGG$\rightarrow$LUAD \\ \midrule
			AttMIL \cite{ilse2018attention}    &2018   &  0.5445$\pm$0.0014    &  0.5561$\pm$0.0033    &   0.5582$\pm$0.0095   &  0.5147$\pm$0.0011    &  0.5159$\pm$0.0011 &      0.5280$\pm$0.0008        \\
    \rowcolor{gray!20}
			CLAM \cite{lu2021data}     &2021   & 0.5429$\pm$0.0076     &    0.5623$\pm$0.0019  & 0.5240$\pm$0.0048     &  0.5321$\pm$0.0031    &  0.5540$\pm$0.0065 &        0.5127$\pm$0.0005    &   \\
			TransMIL \cite{shao2021transmil}  &2021   & 0.5890$\pm$0.0012     & 0.5667$\pm$0.0074     &   0.5100$\pm$0.0005   &   0.5316$\pm$0.0015   & \textcolor{blue}{0.5942$\pm$0.0011}  &   0.5046$\pm$0.0023         &   \\
    \rowcolor{gray!20}
			DSMIL \cite{li2021dual}     &2021   & 0.5816$\pm$0.0037     &   0.5752$\pm$0.0094   &  0.5411$\pm$0.0013    &  0.5428$\pm$0.0021    &  0.5714$\pm$0.0067 &   0.5631$\pm$0.0018         &   \\
			PathOmics \cite{ding2023pathology} &2023   & 0.5861$\pm$0.0072     & 0.5758$\pm$0.0073     &  0.5732$\pm$0.0030    &  0.5490$\pm$0.0021    & 0.5513$\pm$0.0080  &    0.5672$\pm$0.0010        &   \\
    \rowcolor{gray!20}
			CMTA \cite{zhou2023cross}     &2023   & 0.5900$\pm$0.0076     &   0.5806$\pm$0.0041   & \textcolor{blue}{0.5886$\pm$0.0028}     &  0.5699$\pm$0.0014    &  0.5655$\pm$0.0024 &     0.5367$\pm$0.0005       &   \\
			RRTMIL \cite{tang2024feature} &2024   &  0.5714$\pm$0.0031    &   0.5738$\pm$0.0033   &  0.5712$\pm$0.0048    & 0.5511$\pm$0.0006     & 0.5738$\pm$0.0082  &   0.5719$\pm$0.0014        &   \\
    \rowcolor{gray!20}
			MoME \cite{xiong2024mome}     &2024   & 0.5974$\pm$0.0013     &  0.5769$\pm$0.0018    &   0.5862$\pm$0.0047   &  0.5794$\pm$0.0006    & 0.5628$\pm$0.0065  &     0.5872$\pm$0.0031       &   \\
			WiKG \cite{li2024dynamic}     &2024   &  0.5918$\pm$0.0021    &  0.5889$\pm$0.0110    &  0.5772$\pm$0.0007    &  \textcolor{blue}{0.5882$\pm$0.0024}    & 0.5758$\pm$0.0064  &    0.5593$\pm$0.0018        &   \\ 
    \rowcolor{gray!20}
			SurvPath \cite{jaume2024modeling}  &2024   &  \textcolor{blue}{0.6075$\pm$0.0045}    &  \textcolor{blue}{0.5905$\pm$0.0077}    &  0.5799$\pm$0.0019    & 0.5772$\pm$0.0027     & 0.5794$\pm$0.0017  &     \textcolor{blue}{0.5954$\pm$0.0019}       &   \\ 
            DETA (Ours) &-   &   \textcolor{red}{0.6566$\pm$0.0030}         & \textcolor{red}{0.7026$\pm$0.0015}   &   \textcolor{red}{0.6259$\pm$0.0083}   & \textcolor{red}{0.6227$\pm$0.0054}  & \textcolor{red}{0.6528$\pm$0.0039} & \textcolor{red}{0.6452$\pm$0.0033}\\
   \bottomrule \bottomrule

    \rowcolor{gray!20}
			Methods  &Year   & UCEC$\rightarrow$LGG & UCEC$\rightarrow$BLCA & UCEC$\rightarrow$LUAD & LUAD$\rightarrow$LGG & LUAD$\rightarrow$UCEC & LUAD$\rightarrow$BLCA  \\ \midrule
		AttMIL \cite{ilse2018attention}    &2018   &  0.5327$\pm$0.0034    &  0.5238$\pm$0.0047    &  0.5100$\pm$0.0032   &    0.5281$\pm$0.0051  &  0.5409$\pm$0.0120  &   0.5220$\pm$0.0008  \\		 \rowcolor{gray!20}
		CLAM \cite{lu2021data}     &2021   &  0.5178$\pm$0.0021    &  0.5150$\pm$0.0045    &   0.5442$\pm$0.0019   &  0.5525$\pm$0.0083    & 0.5387$\pm$0.0065  &    0.5118$\pm$0.0005  \\
			TransMIL \cite{shao2021transmil}  &2021   & \textcolor{blue}{0.6137$\pm$0.0021}     &  0.5660$\pm$0.0016    &  0.5180$\pm$0.0013    &    0.5345$\pm$0.0027  & 0.5461$\pm$0.0014  &  0.5263$\pm$0.0012 &   \\
    \rowcolor{gray!20}
			DSMIL \cite{li2021dual}     &2021   &   0.5507$\pm$0.0043   &  0.5524$\pm$0.0036    & 0.5079$\pm$0.0011     & 0.5553$\pm$0.0103     & 0.5535$\pm$0.0067  &   0.5387$\pm$0.0039   \\
			PathOmics \cite{ding2023pathology} &2023   &  0.5649$\pm$0.0069    &  0.5780$\pm$0.0013    &  0.5653$\pm$0.0015    &   0.5406$\pm$0.0074   & 0.5609$\pm$0.0023  &   0.5514$\pm$0.0039   \\
    \rowcolor{gray!20}
			CMTA \cite{zhou2023cross}     &2023   &  0.5992$\pm$0.0018    &  0.5589$\pm$0.0008    &  0.5655$\pm$0.0045    &  0.5510$\pm$0.0048    & 0.5650$\pm$0.0094  &   \textcolor{blue}{0.5827$\pm$0.0021}  \\
			RRTMIL \cite{tang2024feature} &2024   &  0.5875$\pm$0.0017    &  0.5524$\pm$0.0008    &  0.5742$\pm$0.0018    & 0.5443$\pm$0.0045     & 0.5611$\pm$0.0008  & 0.5688$\pm$0.0026  &   \\
    \rowcolor{gray!20}
			MoME  \cite{xiong2024mome}    &2024   &  0.5821$\pm$0.0006    & 0.5719$\pm$0.0020     &   0.5711$\pm$0.0014   &  0.5540$\pm$0.0023    & \textcolor{blue}{0.6026$\pm$0.0034}  &   0.5769$\pm$0.0036  \\ 
			WiKG \cite{li2024dynamic}     &2024   &  0.5160$\pm$0.0028    & 0.5404$\pm$0.0025     & 0.5658$\pm$0.0044     &  0.6077$\pm$0.0044    &  0.5641$\pm$0.0113 &   0.5653$\pm$0.0005   \\ 
    \rowcolor{gray!20}
			SurvPath \cite{jaume2024modeling}  &2024   & 0.5973$\pm$0.0014     & \textcolor{blue}{0.5917$\pm$0.0036}     &  \textcolor{blue}{0.5769$\pm$0.0025}    &   \textcolor{blue}{0.6123$\pm$0.0077}   & 0.5714$\pm$0.0037  & 0.5625$\pm$0.0013  \\ 
            DETA (Ours) &-   &  \textcolor{red}{0.6227$\pm$0.0049}    &      \textcolor{red}{0.6245$\pm$0.0069}   &    \textcolor{red}{0.6347$\pm$0.0036}  & \textcolor{red}{0.6198$\pm$0.0061} & \textcolor{red}{0.6426$\pm$0.0018} & \textcolor{red}{0.6127$\pm$0.0108}\\
   \bottomrule
	\end{tabular}}
\end{table*}

\subsection{Adversarial Perturbation for Feature-level Alignment}\label{sec:feat_align}
\zq{Adding perturbation in the source domain is an effective data augmentation strategy for better domain alignment \cite{ahmedpathalign}. An often adopted approach is to add random perturbation to the features of the source domain \cite{luo2023source, yin2023coco}. However, although such randomness is expected to improve the generalization ability of the model, it could ignore key semantic information for alignment.}
Therefore, we introduce adaptively learnable perturbations to more effectively identify \zq{meaningful} perturbation directions, \zq{such that the features of two domains can be more consistent in distribution.}

\zq{Specifically, we add perturbations $\delta^{MP}$ and $\delta^{SP}$ to the node features of the source graph and input them to the MP and SP branch, respectively, and learn the perturbations in an adversarial way with the aid of a domain classifier $D(\cdot)$:
\begin{equation}
\small
	\begin{aligned}
	\min_{\|\delta^{MP}\|\leq\epsilon,\|\delta^{SP}\|\leq\epsilon}\max_{D}~\mathcal{L}_{A{P}}
	&=\mathbb{E}_{G^{t}\in\mathcal{D}^{t}}\log\left(1-D({H}^{t},\hat{{p}}^{t})\right)\\
	&+\mathbb{E}_{G^{s}\in\mathcal{D}^{s}}\log D({H}^{s},\hat{{p}}^{s})
	\end{aligned},
\end{equation}
where ${H}^{s}$ is the output of the graph encoder with the perturbed feature input in source domain, ${H}^{t}$ is the output without perturbation in target domain, and $\hat{{p}}_{i}^{s}$ and $\hat{{p}}_{j}^{t}$ are the category distributions of the two domains predicted by the hazard function. With such a adversarial training process, the feature distributions of both domains can be expected to be better aligned.}



\begin{table*}[htbp]
\footnotesize
	\centering
	\caption{The results of ablation studies in one TCGA datasets as the training set and the other three datasets as the test sets. We highlight the best performing scores in red.}
	\label{tab:LGG}
    \renewcommand{\arraystretch}{0.9}
	\setlength{\tabcolsep}{3.3mm}{
		\begin{tabular}{lccc|cccc}
			\toprule
    \rowcolor{gray!20}
			Methods    & BLCA$\rightarrow$LGG & BLCA$\rightarrow$UCEC & BLCA$\rightarrow$LUAD & LGG$\rightarrow$BLCA & LGG$\rightarrow$UCEC & LGG$\rightarrow$LUAD \\ \midrule
		w/o MP       &  0.5489\textcolor[rgb]{0.0,0.6,0.0}{$_{\downarrow0.1077}$}    &   0.6181\textcolor[rgb]{0.0,0.6,0.0}{$_{\downarrow0.0854}$}   &  0.5790\textcolor[rgb]{0.0,0.6,0.0}{$_{\downarrow0.0469}$}  &  0.5782\textcolor[rgb]{0.0,0.6,0.0}{$_{\downarrow0.0445}$}    &  0.6101\textcolor[rgb]{0.0,0.6,0.0}{$_{\downarrow0.0427}$}  &  0.5699\textcolor[rgb]{0.0,0.6,0.0}{$_{\downarrow0.0753}$}   \\		 \rowcolor{gray!20}
		w/o SP        &  0.5415\textcolor[rgb]{0.0,0.6,0.0}{$_{\downarrow0.1151}$}    &  0.6646\textcolor[rgb]{0.0,0.6,0.0}{$_{\downarrow0.038}$}    &   0.5687\textcolor[rgb]{0.0,0.6,0.0}{$_{\downarrow0.0842}$}   &  0.5687\textcolor[rgb]{0.0,0.6,0.0}{$_{\downarrow0.0540}$}    & 0.5976\textcolor[rgb]{0.0,0.6,0.0}{$_{\downarrow0.0561}$}  &  0.5735\textcolor[rgb]{0.0,0.6,0.0}{$_{\downarrow0.0717}$}    \\
			w/o $\delta^{MP}$     &  0.5368\textcolor[rgb]{0.0,0.6,0.0}{$_{\downarrow0.1198}$}    &  0.6255\textcolor[rgb]{0.0,0.6,0.0}{$_{\downarrow0.0771}$}   &  0.5734\textcolor[rgb]{0.0,0.6,0.0}{$_{\downarrow0.0525}$}         & 0.5584\textcolor[rgb]{0.0,0.6,0.0}{$_{\downarrow0.0643}$}  & 0.6077\textcolor[rgb]{0.0,0.6,0.0}{$_{\downarrow0.0451}$}  & 0.5691\textcolor[rgb]{0.0,0.6,0.0}{$_{\downarrow0.0761}$}   \\
    \rowcolor{gray!20}
			w/o $\delta^{SP}$        &  0.5249\textcolor[rgb]{0.0,0.6,0.0}{$_{\downarrow0.1317}$}    & 0.6134\textcolor[rgb]{0.0,0.6,0.0}{$_{\downarrow0.0892}$}  &  0.5669\textcolor[rgb]{0.0,0.6,0.0}{$_{\downarrow0.0590}$}   &    0.5531\textcolor[rgb]{0.0,0.6,0.0}{$_{\downarrow0.0696}$}  & 0.5929\textcolor[rgb]{0.0,0.6,0.0}{$_{\downarrow0.0599}$}  & 0.5617\textcolor[rgb]{0.0,0.6,0.0}{$_{\downarrow0.0835}$}     \\
			w/o $\delta^{MP}$/$\delta^{SP}$    &  0.5201\textcolor[rgb]{0.0,0.6,0.0}{$_{\downarrow0.1365}$}    &   0.6079\textcolor[rgb]{0.0,0.6,0.0}{$_{\downarrow0.0947}$}   &   0.5618\textcolor[rgb]{0.0,0.6,0.0}{$_{\downarrow0.0641}$}   & 0.5490\textcolor[rgb]{0.0,0.6,0.0}{$_{\downarrow0.0737}$}     & 0.5905\textcolor[rgb]{0.0,0.6,0.0}{$_{\downarrow0.0675}$}  &  0.5600\textcolor[rgb]{0.0,0.6,0.0}{$_{\downarrow0.0852}$}    \\
    \rowcolor{gray!20}
			w/o BC        &   0.5347\textcolor[rgb]{0.0,0.6,0.0}{$_{\downarrow0.1219}$}   & 0.6357\textcolor[rgb]{0.0,0.6,0.0}{$_{\downarrow0.0489}$}    &  0.5721\textcolor[rgb]{0.0,0.6,0.0}{$_{\downarrow0.0538}$}    &  0.5544\textcolor[rgb]{0.0,0.6,0.0}{$_{\downarrow0.0683}$}    &  0.6031\textcolor[rgb]{0.0,0.6,0.0}{$_{\downarrow0.0497}$} &  0.5684\textcolor[rgb]{0.0,0.6,0.0}{$_{\downarrow0.0768}$}   \\
            DETA (Ours)    &   \textcolor{red}{0.6566}         & \textcolor{red}{0.7026}   &   \textcolor{red}{0.6259}   & \textcolor{red}{0.6227}  & \textcolor{red}{0.6528} & \textcolor{red}{0.6452} \\
   \bottomrule    \bottomrule
       \rowcolor{gray!20}
			Methods    & UCEC$\rightarrow$LGG & UCEC$\rightarrow$BLCA & UCEC$\rightarrow$LUAD & LUAD$\rightarrow$LGG & LUAD$\rightarrow$UCEC & LUAD$\rightarrow$BLCA  \\ \midrule
		w/o MP       &  0.5598\textcolor[rgb]{0.0,0.6,0.0}{$_{\downarrow0.0629}$}    &  0.5565\textcolor[rgb]{0.0,0.6,0.0}{$_{\downarrow0.0680}$}    &  0.5721\textcolor[rgb]{0.0,0.6,0.0}{$_{\downarrow0.0626}$}  &   0.5373\textcolor[rgb]{0.0,0.6,0.0}{$_{\downarrow0.0825}$}   &  0.5966\textcolor[rgb]{0.0,0.6,0.0}{$_{\downarrow0.0460}$}  &  0.5864\textcolor[rgb]{0.0,0.6,0.0}{$_{\downarrow0.0263}$}   \\		 \rowcolor{gray!20}
		 w/o SP        &   0.5595\textcolor[rgb]{0.0,0.6,0.0}{$_{\downarrow0.0632}$}   &  0.5491\textcolor[rgb]{0.0,0.6,0.0}{$_{\downarrow0.0754}$}    &   0.5690\textcolor[rgb]{0.0,0.6,0.0}{$_{\downarrow0.0657}$}   &   0.5499\textcolor[rgb]{0.0,0.6,0.0}{$_{\downarrow0.0699}$}   & 0.5993\textcolor[rgb]{0.0,0.6,0.0}{$_{\downarrow0.0433}$}  & 0.5689\textcolor[rgb]{0.0,0.6,0.0}{$_{\downarrow0.0429}$}     \\
			w/o $\delta^{MP}$   & 0.5517\textcolor[rgb]{0.0,0.6,0.0}{$_{\downarrow0.0710}$}     &  0.5556\textcolor[rgb]{0.0,0.6,0.0}{$_{\downarrow0.0689}$} &  0.5718\textcolor[rgb]{0.0,0.6,0.0}{$_{\downarrow0.0656}$}      & 0.5441\textcolor[rgb]{0.0,0.6,0.0}{$_{\downarrow0.0757}$} & 0.5964\textcolor[rgb]{0.0,0.6,0.0}{$_{\downarrow0.0462}$} &  0.5882\textcolor[rgb]{0.0,0.6,0.0}{$_{\downarrow0.0245}$}   \\
    \rowcolor{gray!20}
			w/o $\delta^{SP}$        &  0.5488\textcolor[rgb]{0.0,0.6,0.0}{$_{\downarrow0.0739}$}    & 0.5536\textcolor[rgb]{0.0,0.6,0.0}{$_{\downarrow0.0709}$}  & 0.5711\textcolor[rgb]{0.0,0.6,0.0}{$_{\downarrow0.0636}$}    &   0.5452\textcolor[rgb]{0.0,0.6,0.0}{$_{\downarrow0.0774}$}   & 0.5913\textcolor[rgb]{0.0,0.6,0.0}{$_{\downarrow0.0513}$}  &  0.5689\textcolor[rgb]{0.0,0.6,0.0}{$_{\downarrow0.0438}$}    \\
			w/o $\delta^{MP}$/$\delta^{SP}$    &   0.5412\textcolor[rgb]{0.0,0.6,0.0}{$_{\downarrow0.0815}$}   &  0.5487\textcolor[rgb]{0.0,0.6,0.0}{$_{\downarrow0.0758}$}    &  0.5702\textcolor[rgb]{0.0,0.6,0.0}{$_{\downarrow0.0645}$}    &  0.5359\textcolor[rgb]{0.0,0.6,0.0}{$_{\downarrow0.0839}$}    & 0.5813\textcolor[rgb]{0.0,0.6,0.0}{$_{\downarrow0.0613}$}  &   0.5622\textcolor[rgb]{0.0,0.6,0.0}{$_{\downarrow0.0505}$}   \\
    \rowcolor{gray!20}
			w/o BC        &  0.5459\textcolor[rgb]{0.0,0.6,0.0}{$_{\downarrow0.0768}$}    &  0.5510\textcolor[rgb]{0.0,0.6,0.0}{$_{\downarrow0.0735}$}   &   0.5683\textcolor[rgb]{0.0,0.6,0.0}{$_{\downarrow0.0664}$}   &   0.5397\textcolor[rgb]{0.0,0.6,0.0}{$_{\downarrow0.0801}$}   & 0.5826\textcolor[rgb]{0.0,0.6,0.0}{$_{\downarrow0.0600}$}  &  0.5772\textcolor[rgb]{0.0,0.6,0.0}{$_{\downarrow0.0355}$}   \\
            DETA (Ours)    &  \textcolor{red}{0.6227}    &      \textcolor{red}{0.6245}   &    \textcolor{red}{0.6347}  & \textcolor{red}{0.6198} & \textcolor{red}{0.6426} & \textcolor{red}{0.6127} \\
   \bottomrule
	\end{tabular}}
    \label{tab:abla}
 \vspace{-4mm}
\end{table*}

\subsection{Model Training}
\zq{The proposed method is implemented with a two-stage training. In the pre-training stage, we train the graph encoder $F(\cdot)$ and hazard predictor $h(\cdot)$ on the source domain dataset $\mathcal{D}$ using the survival loss defined in Eq. \eqref{eq:l_sur}. Then in the GDA stage, fine-tune $F(\cdot)$ and $h(\cdot)$ with the combination of the survival loss and the losses proposed in Section \ref{sec:cat_align} and \ref{sec:feat_align}. The overall training objective in the GDA stage can be summarized as follows:}

\begin{equation}
\small
    \min_{{\delta}^{MP},{\delta}^{SP}, h, S}\left\{\mathcal{L}_{surv}+\mathcal{L}_1+\mathcal{L}_2+\max_{D}\{\mathcal{L}_{AP}\}\right\}.
\end{equation}

\section{Experiments}
In this section, we conduct extensive experiments on four TCGA datasets to comprehensively evaluate the effectiveness of our proposed model. First, we describe the datasets and evaluation metrics used in the experiments. Then, we conduct a comprehensive comparison of the experimental results, comparing our model with some of the current state-of-the-art (SOTA) methods. Next, we conduct ablation studies to show the effectiveness of each key component. To further show the effectiveness of our method, we also provide a survival analysis example, and visualize the feature and category alignment effects.

\subsection{Datasets and Evaluation Metrics}
\textbf{Datasets.} The TCGA is a large-scale public database that collects genomic and clinical data of thousands of cancer patients, covering 33 common cancer types \cite{weinstein2013cancer}. TCGA has been widely used in the biomedical research, especially in survival analysis, providing rich data support for identifying genetic variants, and molecular pathways with cancer survival. We selected prognostic data from TCGA to evaluate the performance of our proposed model. The four datasets include: BLCA, LGG, LUAD, and UCEC. We used a 5-fold cross-validation method in each data set, dividing the data into a training set and a validation set to ensure the generalization ability of the model under different samples.

\noindent\textbf{Evaluation Metrics.} C-index is a metric used to evaluate the model's ability to predict survival time ranking. It measures the consistency between the survival time order predicted by the model and the actual order. The value range of C-index is $[0, 1]$. The definition of C-index is as follows:
\begin{equation}
\small
	\text{C-index}=\frac{1}{n(n-1)}\sum_{i=1}^n\sum_{j=1}^nI(T_i<T_j)(1-c_j)
\end{equation}
where $n$ represents the total number of samples, $T_i$ and $T_j$ represent the survival time of the $i$-th patient and the $j$-th patient, respectively, $c_j$ is the censorship status.

\begin{figure*}
    \centering
    \includegraphics[width=1\linewidth]{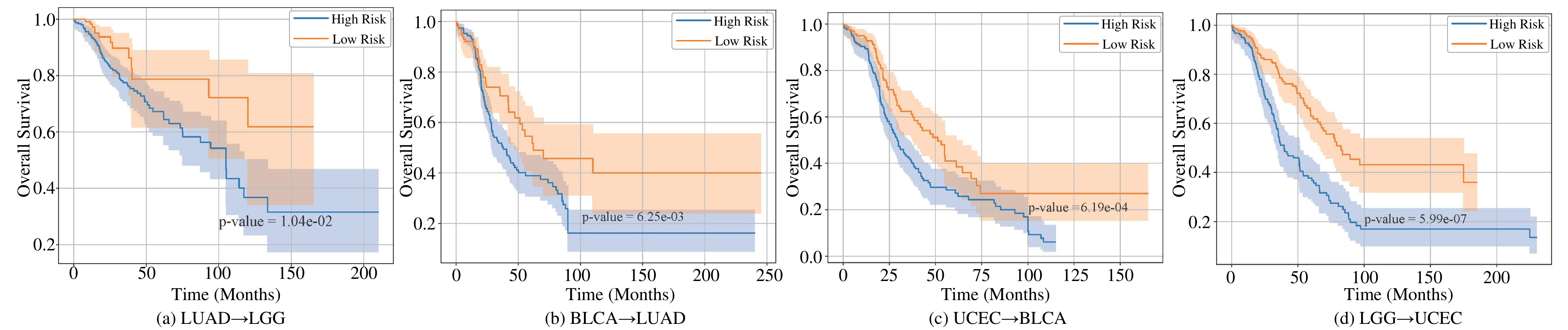}
    \caption{According to the predicted risks, all patients are stratified into low-risk and high-risk group, and Kaplan-Meier analysis is used to display the changes in the survival probability over time. We further use the Log-rank test to perform statistical significance testing.}
    \label{fig.surv}
    \vspace{-3mm}
\end{figure*}


\begin{figure}[htbp]
	\centering
	\subfloat[CMTA]{\includegraphics[width=0.5\linewidth]{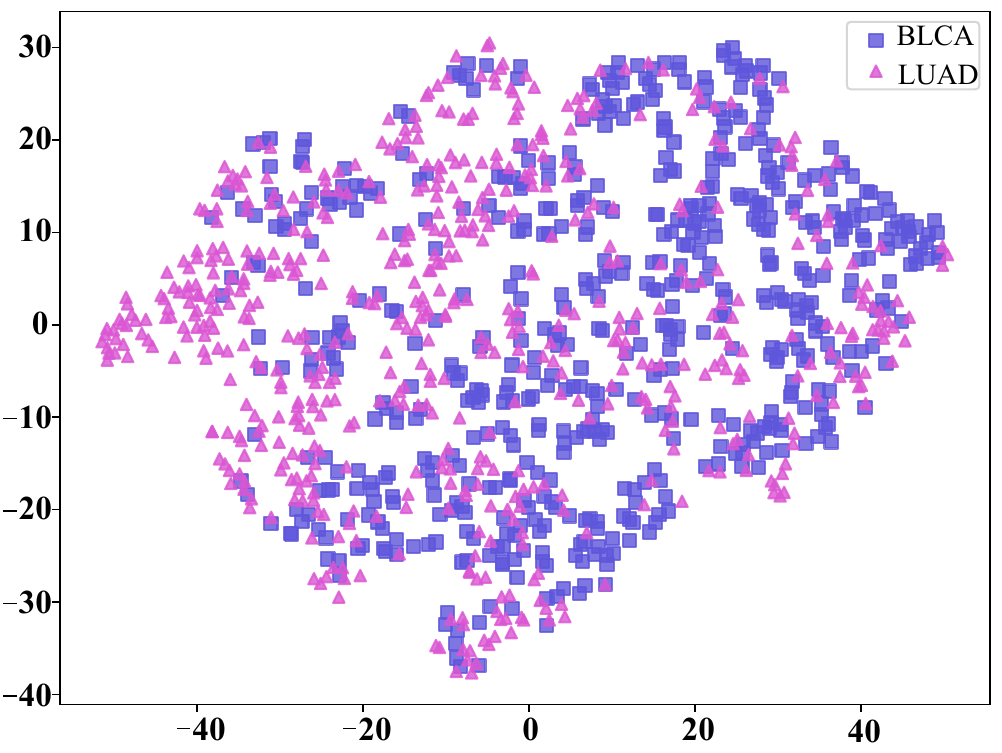}%
		\label{fig:dis1}}
	\hfil
	\subfloat[Ours]{\includegraphics[width=0.5\linewidth]{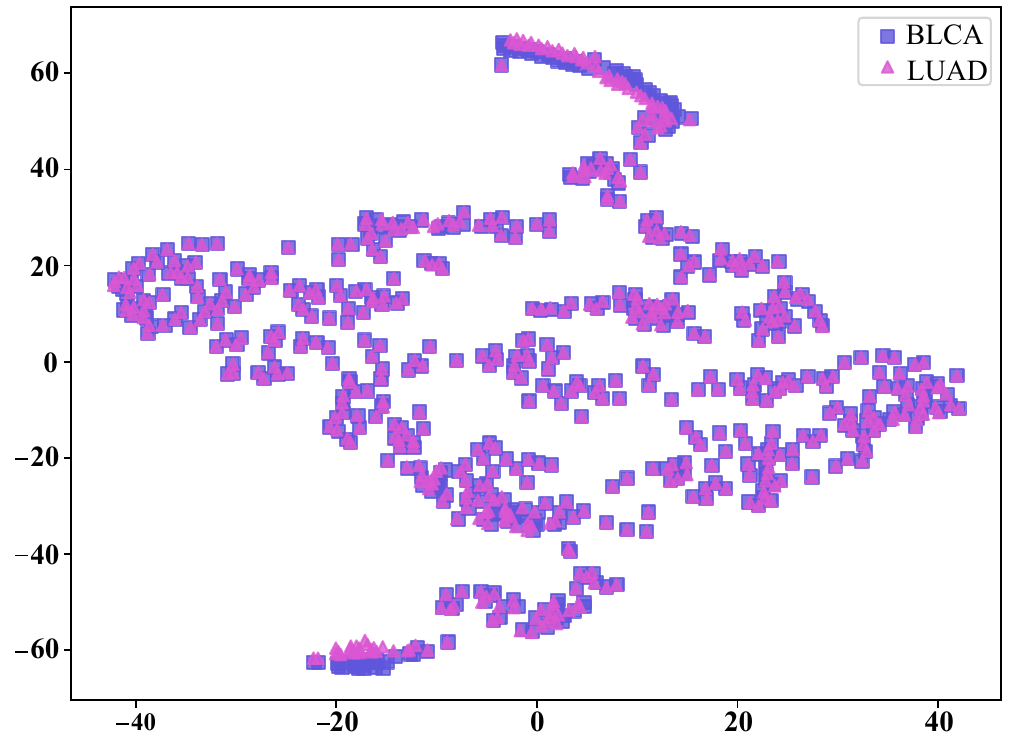}%
		\label{fig:dis2}}
	\caption{t-SNE visualizations of the feature distributions of source and target domains by CMTA and our method.}
	\label{fig:tsne}
	\vspace{-4mm}
\end{figure}

\subsection{Comparisons with SOTA methods}
As shown in Table \ref{tab:BLCA}, our proposed DETA method achieves the best performance on all TCGA datasets. This performance improvement is mainly due to the following two key factors: (1) Using MP branch and SP branch to extract key semantic information from an implicit and explicit perspective respectively significantly enhances the ability to capture global dependencies. (2) DETA further enhances cross-domain category and feature alignment through branch coupling and adversarial perturbation mechanisms. The branch coupling achieves efficient category alignment by fusing the output features of the MP and SP branch. By introducing adversarial perturbations into the source domain features, the model can better align the feature distribution in the target domain, effectively alleviating the distribution difference between the source domain and the target domain.

\subsection{Ablation Studies}
To comprehensively evaluate the impact of each module on the performance of DETA, we conducted ablation experiments. Specifically, we removed different modules one by one and evaluated the model performance. (1) w/o MP: both branches all using the SP branch. (2) w/o SP: both branches all using the MP branch. (3) w/o $\delta^{MP}$: MP branch without perturbation. (4) w/o $\delta^{SP}$: SP branch without perturbation. (5) w/o AP: The perturbation was removed from both the MP branch and the SP branch. (6) w/o BC: The branch coupling was removed.

From Table \ref{tab:abla}, several conclusions can be made. (1) DETA outperforms the variants w/o MP or w/o SP, which highlights the importance of extracting graph semantic information from both implicit and explicit perspectives. The dual-branch design can effectively make up for the shortcomings of a single branch and achieve comprehensive capture of multi-level graph semantics. (2) DETA performs significantly better than the models without perturbation modules (i.e., w/o $\delta^{MP}$, w/o $\delta^{SP}$, and w/o AP). The results show the important role of learnable perturbation modules in cross-domain tasks. (3) w/o BC performs better than w/o AP but worse than DETA. We attribute the fact that although the branch coupling is removed in the w/o BC variant, the perturbation module is still retained, which can alleviate the domain distribution difference. However, due to the lack of category alignment, the model's performance in the target domain is limited.

\subsection{Survival Analysis}
To further verify the effectiveness of the DETA model, we divided all patients into low-risk and high-risk groups based on the risk score predicted by DETA. Specifically, we used the median of the risk score as the boundary of the grouping and used the Kaplan-Meier survival curve analysis method to visualize the survival of all patients. As shown in Fig. \ref{fig.surv}, the analysis results intuitively show the survival differences between different risk groups. To evaluate the statistical significance of the survival difference, we also performed a Logrank test. This test measures whether the survival difference between the low-risk group and the high-risk group is significant by calculating the $p$-value. Generally, a $p$-value less than or equal to 0.05 is considered statistically significant, which means that there is a significant difference in the survival curves between the two groups. It can be clearly seen from the results of Fig. \ref{fig.surv} that the $p$-value is significantly less than 0.05 on all data sets, indicating the potential availability of the proposed method in survival analysis.

\subsection{Visualization of the Feature and Category Alignment}
\begin{figure}
    \centering
    \includegraphics[width=1\linewidth]{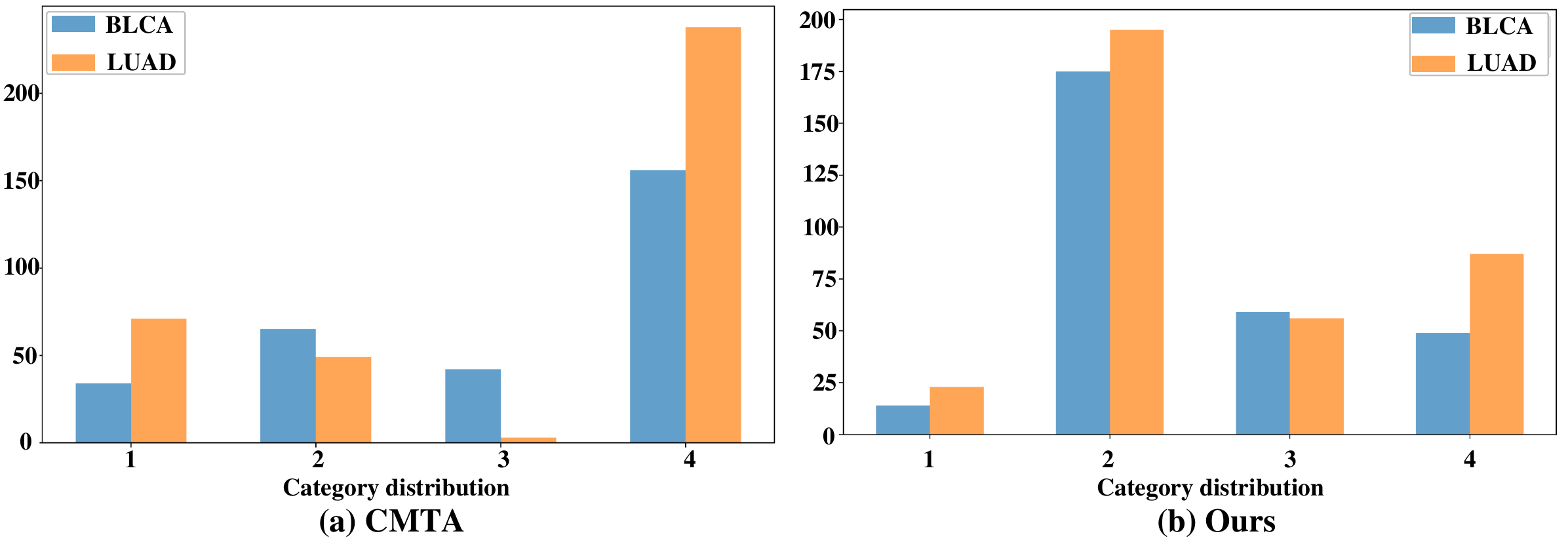}
    \caption{Visualizations of the category distributions of source and target domains by CMTA and our method.}
    \label{fig:dis}
    \vspace{-3mm}
\end{figure}

To better show the mechanism of the proposed method for its promising performance, we visualize the feature and category alignment results in comparison with CMTA, which is one of the SOTA WSI-based survival analysis methods. Specifically, Fig. \ref{fig:tsne} shows the t-SNE embedding of the features extracted from the source and target domains by our model and CMTA, respectively, and we can see a clear feature distribution shift between the source domain and the target domain by CMTA. In contrast, our model minimizes the distribution divergence between the source and target domains.
Furthermore, we also counted the empirical category distributions by both methods in Fig \ref{fig:dis}, and the result shows that the category distributions of the source and target domain by our method look very similar to each other, in comparison with that of CMTA, showing its promising category alignment effect. In summary, the effective alignment at both feature and category levels contributes to the superior GDA performance of our method for the WSI-based survival analysis task.

\section{Conclusions}
In this paper, we propose a DETA framework to explore both feature and category-level alignment between different WSI domains. Specifically, we first formulate the concerned problem as GDA by virtue the graph representation of WSIs. Then we construct a dual-branch graph encoder, including the message passing branch and the shortest path branch, to explicitly and implicitly extract semantic information from the graph-represented WSIs. To realize GDA, we propose a two-level alignment approach: at the category level, we develop a coupling technique by virtue of the dual-branch structure, leading to reduced divergence between the category distributions of the two domains; at the feature level, we introduce an adversarial perturbation strategy to better augment source domain feature, resulting in improved alignment in feature distribution. Extensive experiments on four TCGA datasets have validated the effectiveness of our proposed DETA framework and demonstrated its superior performance in WSI-based survival analysis.

{
    \small
    \bibliographystyle{ieeenat_fullname}
    \bibliography{main}

\begin{thebibliography}{82}
\providecommand{\natexlab}[1]{#1}
\providecommand{\url}[1]{\texttt{#1}}
\expandafter\ifx\csname urlstyle\endcsname\relax
  \providecommand{\doi}[1]{doi: #1}\else
  \providecommand{\doi}{doi: \begingroup \urlstyle{rm}\Url}\fi

\bibitem[Ahmed et~al.()Ahmed, Sellergren, Yang, Xu, Babenko, Ward, Olson,
  Mohtashamian, Matias, Corrado, et~al.]{ahmedpathalign}
Faruk Ahmed, Andrew Sellergren, Lin Yang, Jinhua Xu, Boris Babenko, Abbi Ward,
  Niels Olson, Arash Mohtashamian, Yossi Matias, Greg Corrado, et~al.
\newblock Pathalign: A vision--language model for whole slide images in
  histopathology.
\newblock In \emph{MICCAI Workshop on Computational Pathology with Multimodal
  Data (COMPAYL)}.

\bibitem[Ai et~al.(2023{\natexlab{a}})Ai, Shou, Meng, Yin, and Li]{ai2023gcn}
Wei Ai, Yuntao Shou, Tao Meng, Nan Yin, and Keqin Li.
\newblock Der-gcn: Dialogue and event relation-aware graph convolutional neural
  network for multimodal dialogue emotion recognition.
\newblock \emph{arXiv preprint arXiv:2312.10579}, 2023{\natexlab{a}}.

\bibitem[Ai et~al.(2023{\natexlab{b}})Ai, Zhang, Meng, Shou, Shao, and
  Li]{ai2023two}
Wei Ai, FuChen Zhang, Tao Meng, YunTao Shou, HongEn Shao, and Keqin Li.
\newblock A two-stage multimodal emotion recognition model based on graph
  contrastive learning.
\newblock In \emph{2023 IEEE 29th International Conference on Parallel and
  Distributed Systems (ICPADS)}, pages 397--404. IEEE, 2023{\natexlab{b}}.

\bibitem[Ai et~al.(2024{\natexlab{a}})Ai, Deng, Chen, Du, Meng, and
  Shou]{ai2024mcsff}
Wei Ai, Wen Deng, Hongyi Chen, Jiayi Du, Tao Meng, and Yuntao Shou.
\newblock Mcsff: Multi-modal consistency and specificity fusion framework for
  entity alignment.
\newblock \emph{arXiv preprint arXiv:2410.14584}, 2024{\natexlab{a}}.

\bibitem[Ai et~al.(2024{\natexlab{b}})Ai, Li, Wang, Du, Meng, Shou, and
  Li]{ai2024graph}
Wei Ai, Jianbin Li, Ze Wang, Jiayi Du, Tao Meng, Yuntao Shou, and Keqin Li.
\newblock Graph contrastive learning via cluster-refined negative sampling for
  semi-supervised text classification.
\newblock \emph{arXiv preprint arXiv:2410.18130}, 2024{\natexlab{b}}.

\bibitem[Ai et~al.(2024{\natexlab{c}})Ai, Shou, Meng, and Li]{ai2024gcn}
Wei Ai, Yuntao Shou, Tao Meng, and Keqin Li.
\newblock Der-gcn: Dialog and event relation-aware graph convolutional neural
  network for multimodal dialog emotion recognition.
\newblock \emph{IEEE Transactions on Neural Networks and Learning Systems},
  2024{\natexlab{c}}.

\bibitem[Ai et~al.(2024{\natexlab{d}})Ai, Wei, Shao, Shou, Meng, and
  Li]{ai2024edge}
Wei Ai, Yingying Wei, Hongen Shao, Yuntao Shou, Tao Meng, and Keqin Li.
\newblock Edge-enhanced minimum-margin graph attention network for short text
  classification.
\newblock \emph{Expert Systems with Applications}, 251:\penalty0 124069,
  2024{\natexlab{d}}.

\bibitem[Baroncini et~al.(2019)Baroncini, Zaffaroni, Moiola, Lorefice, Fenu,
  Iaffaldano, Simone, Fanelli, Patti, D’Amico, et~al.]{baroncini2019long}
Damiano Baroncini, Mauro Zaffaroni, Lucia Moiola, Lorena Lorefice, Giuseppe
  Fenu, Pietro Iaffaldano, Marta Simone, Fulvia Fanelli, Francesco Patti,
  Emanuele D’Amico, et~al.
\newblock Long-term follow-up of pediatric ms patients starting treatment with
  injectable first-line agents: a multicentre, italian, retrospective,
  observational study.
\newblock \emph{Multiple Sclerosis Journal}, 25\penalty0 (3):\penalty0
  399--407, 2019.

\bibitem[Borgwardt and Kriegel(2005)]{borgwardt2005shortest}
Karsten~M Borgwardt and Hans-Peter Kriegel.
\newblock Shortest-path kernels on graphs.
\newblock In \emph{Fifth IEEE International Conference on Data Mining
  (ICDM'05)}, pages 8--pp. IEEE, 2005.

\bibitem[Capra et~al.(2017)Capra, Cordioli, Rasia, Gallo, Signori, and
  Sormani]{capra2017assessing}
Ruggero Capra, Cinzia Cordioli, Sarah Rasia, Fabio Gallo, Alessio Signori, and
  Maria~Pia Sormani.
\newblock Assessing long-term prognosis improvement as a consequence of
  treatment pattern changes in ms.
\newblock \emph{Multiple Sclerosis Journal}, 23\penalty0 (13):\penalty0
  1757--1761, 2017.

\bibitem[Chen et~al.(2022)Chen, Chen, Li, Chen, Trister, Krishnan, and
  Mahmood]{chen2022scaling}
Richard~J Chen, Chengkuan Chen, Yicong Li, Tiffany~Y Chen, Andrew~D Trister,
  Rahul~G Krishnan, and Faisal Mahmood.
\newblock Scaling vision transformers to gigapixel images via hierarchical
  self-supervised learning.
\newblock In \emph{Proceedings of the IEEE/CVF Conference on Computer Vision
  and Pattern Recognition}, pages 16144--16155, 2022.

\bibitem[Chen et~al.(2024)Chen, Ding, Lu, Williamson, Jaume, Song, Chen, Zhang,
  Shao, Shaban, et~al.]{chen2024towards}
Richard~J Chen, Tong Ding, Ming~Y Lu, Drew~FK Williamson, Guillaume Jaume,
  Andrew~H Song, Bowen Chen, Andrew Zhang, Daniel Shao, Muhammad Shaban, et~al.
\newblock Towards a general-purpose foundation model for computational
  pathology.
\newblock \emph{Nature Medicine}, 30\penalty0 (3):\penalty0 850--862, 2024.

\bibitem[Dai et~al.(2022)Dai, Wu, Xiao, Shen, and Wang]{dai2022graph}
Quanyu Dai, Xiao-Ming Wu, Jiaren Xiao, Xiao Shen, and Dan Wang.
\newblock Graph transfer learning via adversarial domain adaptation with graph
  convolution.
\newblock \emph{IEEE Transactions on Knowledge and Data Engineering},
  35\penalty0 (5):\penalty0 4908--4922, 2022.

\bibitem[Di et~al.(2022)Di, Zou, Feng, Zhou, Ji, Dai, and
  Gao]{di2022generating}
Donglin Di, Changqing Zou, Yifan Feng, Haiyan Zhou, Rongrong Ji, Qionghai Dai,
  and Yue Gao.
\newblock Generating hypergraph-based high-order representations of whole-slide
  histopathological images for survival prediction.
\newblock \emph{IEEE Transactions on Pattern Analysis and Machine
  Intelligence}, 45\penalty0 (5):\penalty0 5800--5815, 2022.

\bibitem[Ding et~al.(2023)Ding, Zhou, Metaxas, and Zhang]{ding2023pathology}
Kexin Ding, Mu Zhou, Dimitris~N Metaxas, and Shaoting Zhang.
\newblock Pathology-and-genomics multimodal transformer for survival outcome
  prediction.
\newblock In \emph{International Conference on Medical Image Computing and
  Computer-Assisted Intervention}, pages 622--631. Springer, 2023.

\bibitem[Ding et~al.()Ding, Kong, Chen, Kirchenbauer, Goldblum, Wipf, Huang,
  and Goldstein]{ding2021closer}
Mucong Ding, Kezhi Kong, Jiuhai Chen, John Kirchenbauer, Micah Goldblum, David
  Wipf, Furong Huang, and Tom Goldstein.
\newblock A closer look at distribution shifts and out-of-distribution
  generalization on graphs.
\newblock In \emph{NeurIPS 2021 Workshop on Distribution Shifts: Connecting
  Methods and Applications}.

\bibitem[Feng et~al.(2023)Feng, Li, Zhang, and ZHOU]{fengtowards}
Kaituo Feng, Changsheng Li, Xiaolu Zhang, and JUN ZHOU.
\newblock Towards open temporal graph neural networks.
\newblock In \emph{The Eleventh International Conference on Learning
  Representations}, 2023.

\bibitem[Francone et~al.(2020)Francone, Iafrate, Masci, Coco, Cilia, Manganaro,
  Panebianco, Andreoli, Colaiacomo, Zingaropoli, et~al.]{francone2020chest}
Marco Francone, Franco Iafrate, Giorgio~Maria Masci, Simona Coco, Francesco
  Cilia, Lucia Manganaro, Valeria Panebianco, Chiara Andreoli, Maria~Chiara
  Colaiacomo, Maria~Antonella Zingaropoli, et~al.
\newblock Chest ct score in covid-19 patients: correlation with disease
  severity and short-term prognosis.
\newblock \emph{European radiology}, 30:\penalty0 6808--6817, 2020.

\bibitem[Gao et~al.(2021)Gao, Liu, and Ji]{gao2021topology}
Hongyang Gao, Yi Liu, and Shuiwang Ji.
\newblock Topology-aware graph pooling networks.
\newblock \emph{IEEE Transactions on Pattern Analysis and Machine
  Intelligence}, 43\penalty0 (12):\penalty0 4512--4518, 2021.

\bibitem[Guo et~al.(2022)Guo, Wang, Yan, Lou, Feng, Zhu, Chen, He, and
  Philip]{guo2022learning}
Gaoyang Guo, Chaokun Wang, Bencheng Yan, Yunkai Lou, Hao Feng, Junchao Zhu, Jun
  Chen, Fei He, and S~Yu Philip.
\newblock Learning adaptive node embeddings across graphs.
\newblock \emph{IEEE Transactions on Knowledge and Data Engineering},
  35\penalty0 (6):\penalty0 6028--6042, 2022.

\bibitem[Ilse et~al.(2018)Ilse, Tomczak, and Welling]{ilse2018attention}
Maximilian Ilse, Jakub Tomczak, and Max Welling.
\newblock Attention-based deep multiple instance learning.
\newblock In \emph{International conference on machine learning}, pages
  2127--2136. PMLR, 2018.

\bibitem[Inusah et~al.(2010)Inusah, Sormani, Cofield, Aban, Musani,
  Srinivasasainagendra, and Cutter]{inusah2010assessing}
Seidu Inusah, Maria~P Sormani, Stacey~S Cofield, Inmaculada~B Aban, Solomon~K
  Musani, Vinodh Srinivasasainagendra, and Gary~R Cutter.
\newblock Assessing changes in relapse rates in multiple sclerosis.
\newblock \emph{Multiple sclerosis journal}, 16\penalty0 (12):\penalty0
  1414--1421, 2010.

\bibitem[Jaume et~al.(2024{\natexlab{a}})Jaume, Oldenburg, Vaidya, Chen,
  Williamson, Peeters, Song, and Mahmood]{jaume2024transcriptomics}
Guillaume Jaume, Lukas Oldenburg, Anurag Vaidya, Richard~J Chen, Drew~FK
  Williamson, Thomas Peeters, Andrew~H Song, and Faisal Mahmood.
\newblock Transcriptomics-guided slide representation learning in computational
  pathology.
\newblock In \emph{Proceedings of the IEEE/CVF Conference on Computer Vision
  and Pattern Recognition}, pages 9632--9644, 2024{\natexlab{a}}.

\bibitem[Jaume et~al.(2024{\natexlab{b}})Jaume, Vaidya, Chen, Williamson,
  Liang, and Mahmood]{jaume2024modeling}
Guillaume Jaume, Anurag Vaidya, Richard~J Chen, Drew~FK Williamson, Paul~Pu
  Liang, and Faisal Mahmood.
\newblock Modeling dense multimodal interactions between biological pathways
  and histology for survival prediction.
\newblock In \emph{Proceedings of the IEEE/CVF Conference on Computer Vision
  and Pattern Recognition}, pages 11579--11590, 2024{\natexlab{b}}.

\bibitem[Kalafi et~al.(2019)Kalafi, Nor, Taib, Ganggayah, Town, and
  Dhillon]{kalafi2019machine}
EY Kalafi, NAM Nor, NA Taib, MD Ganggayah, C Town, and SK Dhillon.
\newblock Machine learning and deep learning approaches in breast cancer
  survival prediction using clinical data.
\newblock \emph{Folia biologica}, 65\penalty0 (5-6):\penalty0 212--220, 2019.

\bibitem[Li et~al.(2021)Li, Li, and Eliceiri]{li2021dual}
Bin Li, Yin Li, and Kevin~W Eliceiri.
\newblock Dual-stream multiple instance learning network for whole slide image
  classification with self-supervised contrastive learning.
\newblock In \emph{Proceedings of the IEEE/CVF conference on computer vision
  and pattern recognition}, pages 14318--14328, 2021.

\bibitem[Li et~al.(2024)Li, Chen, Chu, Sun, Guan, Han, and He]{li2024dynamic}
Jiawen Li, Yuxuan Chen, Hongbo Chu, Qiehe Sun, Tian Guan, Anjia Han, and
  Yonghong He.
\newblock Dynamic graph representation with knowledge-aware attention for
  histopathology whole slide image analysis.
\newblock In \emph{Proceedings of the IEEE/CVF Conference on Computer Vision
  and Pattern Recognition}, pages 11323--11332, 2024.

\bibitem[Li et~al.(2018)Li, Yao, Zhu, Li, and Huang]{li2018graph}
Ruoyu Li, Jiawen Yao, Xinliang Zhu, Yeqing Li, and Junzhou Huang.
\newblock Graph cnn for survival analysis on whole slide pathological images.
\newblock In \emph{International Conference on Medical Image Computing and
  Computer-Assisted Intervention}, pages 174--182. Springer, 2018.

\bibitem[Lin et~al.(2023)Lin, Li, Li, Chen, Li, and Lu]{lin2023multi}
Mingkai Lin, Wenzhong Li, Ding Li, Yizhou Chen, Guohao Li, and Sanglu Lu.
\newblock Multi-domain generalized graph meta learning.
\newblock In \emph{Proceedings of the AAAI Conference on Artificial
  Intelligence}, pages 4479--4487, 2023.

\bibitem[Lu et~al.(2021)Lu, Williamson, Chen, Chen, Barbieri, and
  Mahmood]{lu2021data}
Ming~Y Lu, Drew~FK Williamson, Tiffany~Y Chen, Richard~J Chen, Matteo Barbieri,
  and Faisal Mahmood.
\newblock Data-efficient and weakly supervised computational pathology on
  whole-slide images.
\newblock \emph{Nature biomedical engineering}, 5\penalty0 (6):\penalty0
  555--570, 2021.

\bibitem[Lu et~al.(2023)Lu, Chen, Zhang, Williamson, Chen, Ding, Le, Chuang,
  and Mahmood]{lu2023visual}
Ming~Y Lu, Bowen Chen, Andrew Zhang, Drew~FK Williamson, Richard~J Chen, Tong
  Ding, Long~Phi Le, Yung-Sung Chuang, and Faisal Mahmood.
\newblock Visual language pretrained multiple instance zero-shot transfer for
  histopathology images.
\newblock In \emph{Proceedings of the IEEE/CVF conference on Computer Vision
  and Pattern Recognition}, pages 19764--19775, 2023.

\bibitem[Luo et~al.(2023)Luo, Wang, Chen, Huang, and
  Baktashmotlagh]{luo2023source}
Yadan Luo, Zijian Wang, Zhuoxiao Chen, Zi Huang, and Mahsa Baktashmotlagh.
\newblock Source-free progressive graph learning for open-set domain
  adaptation.
\newblock \emph{IEEE Transactions on Pattern Analysis and Machine
  Intelligence}, 45\penalty0 (9):\penalty0 11240--11255, 2023.

\bibitem[Ma et~al.(2021)Ma, Gao, and Xu]{ma2021active}
Xinhong Ma, Junyu Gao, and Changsheng Xu.
\newblock Active universal domain adaptation.
\newblock In \emph{Proceedings of the IEEE/CVF International Conference on
  Computer Vision}, pages 8968--8977, 2021.

\bibitem[Mancini et~al.(2019)Mancini, Porzi, Bulo, Caputo, and
  Ricci]{mancini2019inferring}
Massimiliano Mancini, Lorenzo Porzi, Samuel~Rota Bulo, Barbara Caputo, and
  Elisa Ricci.
\newblock Inferring latent domains for unsupervised deep domain adaptation.
\newblock \emph{IEEE Transactions on Pattern Analysis and Machine
  Intelligence}, 43\penalty0 (2):\penalty0 485--498, 2019.

\bibitem[Meng et~al.(2024{\natexlab{a}})Meng, Shou, Ai, Du, Liu, and
  Li]{meng2024multi}
Tao Meng, Yuntao Shou, Wei Ai, Jiayi Du, Haiyan Liu, and Keqin Li.
\newblock A multi-message passing framework based on heterogeneous graphs in
  conversational emotion recognition.
\newblock \emph{Neurocomputing}, 569:\penalty0 127109, 2024{\natexlab{a}}.

\bibitem[Meng et~al.(2024{\natexlab{b}})Meng, Shou, Ai, Yin, and
  Li]{meng2024deep}
Tao Meng, Yuntao Shou, Wei Ai, Nan Yin, and Keqin Li.
\newblock Deep imbalanced learning for multimodal emotion recognition in
  conversations.
\newblock \emph{IEEE Transactions on Artificial Intelligence},
  2024{\natexlab{b}}.

\bibitem[Meng et~al.(2024{\natexlab{c}})Meng, Zhang, Shou, Ai, Yin, and
  Li]{meng2024revisiting}
Tao Meng, Fuchen Zhang, Yuntao Shou, Wei Ai, Nan Yin, and Keqin Li.
\newblock Revisiting multimodal emotion recognition in conversation from the
  perspective of graph spectrum.
\newblock \emph{arXiv preprint arXiv:2404.17862}, 2024{\natexlab{c}}.

\bibitem[Meng et~al.(2024{\natexlab{d}})Meng, Zhang, Shou, Shao, Ai, and
  Li]{meng2024masked}
Tao Meng, Fuchen Zhang, Yuntao Shou, Hongen Shao, Wei Ai, and Keqin Li.
\newblock Masked graph learning with recurrent alignment for multimodal emotion
  recognition in conversation.
\newblock \emph{IEEE/ACM Transactions on Audio, Speech, and Language
  Processing}, 2024{\natexlab{d}}.

\bibitem[Nakhli et~al.(2023)Nakhli, Zhang, Mirabadi, Rich, Asadi, Gilks,
  Farahani, and Bashashati]{nakhli2023co}
Ramin Nakhli, Allen Zhang, Ali Mirabadi, Katherine Rich, Maryam Asadi, Blake
  Gilks, Hossein Farahani, and Ali Bashashati.
\newblock Co-pilot: Dynamic top-down point cloud with conditional neighborhood
  aggregation for multi-gigapixel histopathology image representation.
\newblock In \emph{Proceedings of the IEEE/CVF International Conference on
  Computer Vision}, pages 21063--21073, 2023.

\bibitem[Platz et~al.(2017)Platz, Merz, Jhund, Vazir, Campbell, and
  McMurray]{platz2017dynamic}
Elke Platz, Allison~A Merz, Pardeep~S Jhund, Ali Vazir, Ross Campbell, and
  John~J McMurray.
\newblock Dynamic changes and prognostic value of pulmonary congestion by lung
  ultrasound in acute and chronic heart failure: a systematic review.
\newblock \emph{European journal of heart failure}, 19\penalty0 (9):\penalty0
  1154--1163, 2017.

\bibitem[Ruggieri et~al.(2024)Ruggieri, Prosperini, Al-Araji, Annovazzi,
  Bisecco, Ciccarelli, De~Stefano, Filippi, Fleischer, Evangelou,
  et~al.]{ruggieri2024assessing}
Serena Ruggieri, Luca Prosperini, Sarmad Al-Araji, Pietro~Osvaldo Annovazzi,
  Alvino Bisecco, Olga Ciccarelli, Nicola De~Stefano, Massimo Filippi, Vinzenz
  Fleischer, Nikos Evangelou, et~al.
\newblock Assessing treatment response to oral drugs for multiple sclerosis in
  real-world setting: a magnims study.
\newblock \emph{Journal of Neurology, Neurosurgery \& Psychiatry}, 95\penalty0
  (2):\penalty0 142--150, 2024.

\bibitem[Saito et~al.(2018)Saito, Watanabe, Ushiku, and
  Harada]{saito2018maximum}
Kuniaki Saito, Kohei Watanabe, Yoshitaka Ushiku, and Tatsuya Harada.
\newblock Maximum classifier discrepancy for unsupervised domain adaptation.
\newblock In \emph{Proceedings of the IEEE Conference on Computer Vision and
  Pattern Recognition}, pages 3723--3732, 2018.

\bibitem[Shao et~al.(2021)Shao, Bian, Chen, Wang, Zhang, Ji,
  et~al.]{shao2021transmil}
Zhuchen Shao, Hao Bian, Yang Chen, Yifeng Wang, Jian Zhang, Xiangyang Ji,
  et~al.
\newblock Transmil: Transformer based correlated multiple instance learning for
  whole slide image classification.
\newblock \emph{Advances in neural information processing systems},
  34:\penalty0 2136--2147, 2021.

\bibitem[Shen et~al.(2018)Shen, Qu, Zhang, and Yu]{shen2018wasserstein}
Jian Shen, Yanru Qu, Weinan Zhang, and Yong Yu.
\newblock Wasserstein distance guided representation learning for domain
  adaptation.
\newblock In \emph{Proceedings of the AAAI conference on artificial
  intelligence}, 2018.

\bibitem[Shou et~al.(2022{\natexlab{a}})Shou, Meng, Ai, Xie, Liu, and
  Wang]{shou2022object}
Yuntao Shou, Tao Meng, Wei Ai, Canhao Xie, Haiyan Liu, and Yina Wang.
\newblock Object detection in medical images based on hierarchical transformer
  and mask mechanism.
\newblock \emph{Computational Intelligence and Neuroscience}, 2022\penalty0
  (1):\penalty0 5863782, 2022{\natexlab{a}}.

\bibitem[Shou et~al.(2022{\natexlab{b}})Shou, Meng, Ai, Yang, and
  Li]{shou2022conversational}
Yuntao Shou, Tao Meng, Wei Ai, Sihan Yang, and Keqin Li.
\newblock Conversational emotion recognition studies based on graph
  convolutional neural networks and a dependent syntactic analysis.
\newblock \emph{Neurocomputing}, 501:\penalty0 629--639, 2022{\natexlab{b}}.

\bibitem[Shou et~al.(2023{\natexlab{a}})Shou, Ai, Meng, and Li]{shou2023czl}
Yuntao Shou, Wei Ai, Tao Meng, and Keqin Li.
\newblock Czl-ciae: Clip-driven zero-shot learning for correcting inverse age
  estimation.
\newblock \emph{arXiv preprint arXiv:2312.01758}, 2023{\natexlab{a}}.

\bibitem[Shou et~al.(2023{\natexlab{b}})Shou, Ai, Meng, and Yin]{shou2023graph}
Yuntao Shou, Wei Ai, Tao Meng, and Nan Yin.
\newblock Graph information bottleneck for remote sensing segmentation.
\newblock \emph{arXiv preprint arXiv:2312.02545}, 2023{\natexlab{b}}.

\bibitem[Shou et~al.(2023{\natexlab{c}})Shou, Ai, Meng, Zhang, and
  Li]{shou2023graphunet}
YunTao Shou, Wei Ai, Tao Meng, FuChen Zhang, and KeQin Li.
\newblock Graphunet: Graph make strong encoders for remote sensing
  segmentation.
\newblock In \emph{2023 IEEE 29th International Conference on Parallel and
  Distributed Systems (ICPADS)}, pages 2734--2737. IEEE, 2023{\natexlab{c}}.

\bibitem[Shou et~al.(2023{\natexlab{d}})Shou, Meng, Ai, Yin, and
  Li]{shou2023adversarial}
Yuntao Shou, Tao Meng, Wei Ai, Nan Yin, and Keqin Li.
\newblock Adversarial representation with intra-modal and inter-modal graph
  contrastive learning for multimodal emotion recognition.
\newblock \emph{arXiv preprint arXiv:2312.16778}, 2023{\natexlab{d}}.

\bibitem[Shou et~al.(2023{\natexlab{e}})Shou, Meng, Ai, Yin, and
  Li]{shou2023comprehensive}
Yuntao Shou, Tao Meng, Wei Ai, Nan Yin, and Keqin Li.
\newblock A comprehensive survey on multi-modal conversational emotion
  recognition with deep learning.
\newblock \emph{arXiv preprint arXiv:2312.05735}, 2023{\natexlab{e}}.

\bibitem[Shou et~al.(2024{\natexlab{a}})Shou, Ai, Du, Meng, and
  Liu]{shou2024efficient}
Yuntao Shou, Wei Ai, Jiayi Du, Tao Meng, and Haiyan Liu.
\newblock Efficient long-distance latent relation-aware graph neural network
  for multi-modal emotion recognition in conversations.
\newblock \emph{arXiv preprint arXiv:2407.00119}, 2024{\natexlab{a}}.

\bibitem[Shou et~al.(2024{\natexlab{b}})Shou, Cao, and Meng]{shou2024spegcl}
Yuntao Shou, Xiangyong Cao, and Deyu Meng.
\newblock Spegcl: Self-supervised graph spectrum contrastive learning without
  positive samples.
\newblock \emph{arXiv preprint arXiv:2410.10365}, 2024{\natexlab{b}}.

\bibitem[Shou et~al.(2024{\natexlab{c}})Shou, Lan, and
  Cao]{shou2024contrastive}
Yuntao Shou, Haozhi Lan, and Xiangyong Cao.
\newblock Contrastive graph representation learning with adversarial cross-view
  reconstruction and information bottleneck.
\newblock \emph{arXiv preprint arXiv:2408.00295}, 2024{\natexlab{c}}.

\bibitem[Shou et~al.(2024{\natexlab{d}})Shou, Liu, Cao, Meng, and
  Dong]{shou2024low}
Yuntao Shou, Huan Liu, Xiangyong Cao, Deyu Meng, and Bo Dong.
\newblock A low-rank matching attention based cross-modal feature fusion method
  for conversational emotion recognition.
\newblock \emph{IEEE Transactions on Affective Computing}, 2024{\natexlab{d}}.

\bibitem[Shou et~al.(2024{\natexlab{e}})Shou, Meng, Ai, Zhang, Yin, and
  Li]{shou2024adversarial}
Yuntao Shou, Tao Meng, Wei Ai, Fuchen Zhang, Nan Yin, and Keqin Li.
\newblock Adversarial alignment and graph fusion via information bottleneck for
  multimodal emotion recognition in conversations.
\newblock \emph{Information Fusion}, 112:\penalty0 102590, 2024{\natexlab{e}}.

\bibitem[Shou et~al.(2024{\natexlab{f}})Shou, Meng, Zhang, Yin, and
  Li]{shou2024revisiting}
Yuntao Shou, Tao Meng, Fuchen Zhang, Nan Yin, and Keqin Li.
\newblock Revisiting multi-modal emotion learning with broad state space models
  and probability-guidance fusion.
\newblock \emph{arXiv preprint arXiv:2404.17858}, 2024{\natexlab{f}}.

\bibitem[Shou et~al.(2025)Shou, Cao, Liu, and Meng]{shou2025masked}
Yuntao Shou, Xiangyong Cao, Huan Liu, and Deyu Meng.
\newblock Masked contrastive graph representation learning for age estimation.
\newblock \emph{Pattern Recognition}, 158:\penalty0 110974, 2025.

\bibitem[Singh(2021)]{singh2021clda}
Ankit Singh.
\newblock Clda: Contrastive learning for semi-supervised domain adaptation.
\newblock \emph{Advances in Neural Information Processing Systems},
  34:\penalty0 5089--5101, 2021.

\bibitem[Sormani and Bruzzi(2015)]{sormani2015can}
Maria~Pia Sormani and Paolo Bruzzi.
\newblock Can we measure long-term treatment effects in multiple sclerosis?
\newblock \emph{Nature Reviews Neurology}, 11\penalty0 (3):\penalty0 176--182,
  2015.

\bibitem[Srinidhi et~al.(2021)Srinidhi, Ciga, and Martel]{srinidhi2021deep}
Chetan~L Srinidhi, Ozan Ciga, and Anne~L Martel.
\newblock Deep neural network models for computational histopathology: A
  survey.
\newblock \emph{Medical image analysis}, 67:\penalty0 101813, 2021.

\bibitem[Sun et~al.(2022)Sun, Zhou, He, Wang, and Wang]{sun2022gppt}
Mingchen Sun, Kaixiong Zhou, Xin He, Ying Wang, and Xin Wang.
\newblock Gppt: Graph pre-training and prompt tuning to generalize graph neural
  networks.
\newblock In \emph{Proceedings of the 28th ACM SIGKDD Conference on Knowledge
  Discovery and Data Mining}, pages 1717--1727, 2022.

\bibitem[Tang et~al.(2024)Tang, Zhou, Huang, Zhu, Zhang, and
  Liu]{tang2024feature}
Wenhao Tang, Fengtao Zhou, Sheng Huang, Xiang Zhu, Yi Zhang, and Bo Liu.
\newblock Feature re-embedding: Towards foundation model-level performance in
  computational pathology.
\newblock In \emph{Proceedings of the IEEE/CVF Conference on Computer Vision
  and Pattern Recognition}, pages 11343--11352, 2024.

\bibitem[Vale-Silva and Rohr(2021)]{vale2021long}
Lu{\'\i}s~A Vale-Silva and Karl Rohr.
\newblock Long-term cancer survival prediction using multimodal deep learning.
\newblock \emph{Scientific Reports}, 11\penalty0 (1):\penalty0 13505, 2021.

\bibitem[Wang et~al.(2024)Wang, Ma, Gao, Bain, Imoto, Li{\`o}, Cai, Chen, and
  Song]{wang2024dual}
Zhikang Wang, Jiani Ma, Qian Gao, Chris Bain, Seiya Imoto, Pietro Li{\`o},
  Hongmin Cai, Hao Chen, and Jiangning Song.
\newblock Dual-stream multi-dependency graph neural network enables precise
  cancer survival analysis.
\newblock \emph{Medical Image Analysis}, 97:\penalty0 103252, 2024.

\bibitem[Weinstein et~al.(2013)Weinstein, Collisson, Mills, Shaw, Ozenberger,
  Ellrott, Shmulevich, Sander, and Stuart]{weinstein2013cancer}
John~N Weinstein, Eric~A Collisson, Gordon~B Mills, Kenna~R Shaw, Brad~A
  Ozenberger, Kyle Ellrott, Ilya Shmulevich, Chris Sander, and Joshua~M Stuart.
\newblock The cancer genome atlas pan-cancer analysis project.
\newblock \emph{Nature genetics}, 45\penalty0 (10):\penalty0 1113--1120, 2013.

\bibitem[Wu et~al.(2022{\natexlab{a}})Wu, Pan, and Zhu]{wu2022attraction}
Man Wu, Shirui Pan, and Xingquan Zhu.
\newblock Attraction and repulsion: Unsupervised domain adaptive graph
  contrastive learning network.
\newblock \emph{IEEE Transactions on Emerging Topics in Computational
  Intelligence}, 6\penalty0 (5):\penalty0 1079--1091, 2022{\natexlab{a}}.

\bibitem[Wu et~al.(2022{\natexlab{b}})Wu, Zhang, Yan, and Wipf]{wuhandling}
Qitian Wu, Hengrui Zhang, Junchi Yan, and David Wipf.
\newblock Handling distribution shifts on graphs: An invariance perspective.
\newblock In \emph{International Conference on Learning Representations},
  2022{\natexlab{b}}.

\bibitem[Xiong et~al.(2024)Xiong, Chen, Zheng, Wei, Zheng, Sung, and
  King]{xiong2024mome}
Conghao Xiong, Hao Chen, Hao Zheng, Dong Wei, Yefeng Zheng, Joseph~JY Sung, and
  Irwin King.
\newblock Mome: Mixture of multimodal experts for cancer survival prediction.
\newblock In \emph{International Conference on Medical Image Computing and
  Computer-Assisted Intervention}, pages 318--328. Springer, 2024.

\bibitem[Yang et~al.(2020)Yang, Deng, Liu, and Tao]{yang2020heterogeneous}
Xu Yang, Cheng Deng, Tongliang Liu, and Dacheng Tao.
\newblock Heterogeneous graph attention network for unsupervised
  multiple-target domain adaptation.
\newblock \emph{IEEE Transactions on Pattern Analysis and Machine
  Intelligence}, 44\penalty0 (4):\penalty0 1992--2003, 2020.

\bibitem[Yehudai et~al.(2021)Yehudai, Fetaya, Meirom, Chechik, and
  Maron]{yehudai2021local}
Gilad Yehudai, Ethan Fetaya, Eli Meirom, Gal Chechik, and Haggai Maron.
\newblock From local structures to size generalization in graph neural
  networks.
\newblock In \emph{International Conference on Machine Learning}, pages
  11975--11986. PMLR, 2021.

\bibitem[Yin et~al.(2023)Yin, Shen, Wang, Lan, Ma, Chen, Hua, and
  Luo]{yin2023coco}
Nan Yin, Li Shen, Mengzhu Wang, Long Lan, Zeyu Ma, Chong Chen, Xian-Sheng Hua,
  and Xiao Luo.
\newblock Coco: A coupled contrastive framework for unsupervised domain
  adaptive graph classification.
\newblock In \emph{International Conference on Machine Learning}, pages
  40040--40053. PMLR, 2023.

\bibitem[Yin et~al.(2024)Yin, Wang, Chen, Shen, Xiong, Gu, and
  Luo]{yin2024dream}
Nan Yin, Mengzhu Wang, Zhenghan Chen, Li Shen, Huan Xiong, Bin Gu, and Xiao
  Luo.
\newblock Dream: Dual structured exploration with mixup for open-set graph
  domain adaption.
\newblock In \emph{The Twelfth International Conference on Learning
  Representations}, 2024.

\bibitem[Ying et~al.(2021)Ying, Shou, and Liu]{ying2021prediction}
RunKai Ying, Yuntao Shou, and Chang Liu.
\newblock Prediction model of dow jones index based on lstm-adaboost.
\newblock In \emph{2021 International Conference on Communications, Information
  System and Computer Engineering (CISCE)}, pages 808--812. IEEE, 2021.

\bibitem[Yperman et~al.(2022)Yperman, Popescu, Van~Wijmeersch, Becker, and
  Peeters]{yperman2022motor}
Jan Yperman, Veronica Popescu, Bart Van~Wijmeersch, Thijs Becker, and Liesbet~M
  Peeters.
\newblock Motor evoked potentials for multiple sclerosis, a multiyear follow-up
  dataset.
\newblock \emph{Scientific Data}, 9\penalty0 (1):\penalty0 207, 2022.

\bibitem[Yu et~al.(2021)Yu, Zhang, Cao, Xu, You, Chen, Zhu, Kong, Song, Xin,
  et~al.]{yu2021dynamic}
Zhenjun Yu, Yu Zhang, Yingying Cao, Manman Xu, Shaoli You, Yu Chen, Bing Zhu,
  Ming Kong, Fangjiao Song, Shaojie Xin, et~al.
\newblock A dynamic prediction model for prognosis of acute-on-chronic liver
  failure based on the trend of clinical indicators.
\newblock \emph{Scientific reports}, 11\penalty0 (1):\penalty0 1810, 2021.

\bibitem[Zhang et~al.(2024)Zhang, Shou, Meng, Ai, and Li]{zhang2024multi}
Yiping Zhang, Yuntao Shou, Tao Meng, Wei Ai, and Keqin Li.
\newblock A multi-view mask contrastive learning graph convolutional neural
  network for age estimation.
\newblock \emph{Knowledge and Information Systems}, pages 1--26, 2024.

\bibitem[Zhao et~al.(2020)Zhao, Yang, Fang, Liu, Zhou, Zhang, Sun, Yang, Menze,
  Fan, et~al.]{zhao2020predicting}
Yu Zhao, Fan Yang, Yuqi Fang, Hailing Liu, Niyun Zhou, Jun Zhang, Jiarui Sun,
  Sen Yang, Bjoern Menze, Xinjuan Fan, et~al.
\newblock Predicting lymph node metastasis using histopathological images based
  on multiple instance learning with deep graph convolution.
\newblock In \emph{Proceedings of the IEEE/CVF Conference on Computer Vision
  and Pattern Recognition}, pages 4837--4846, 2020.

\bibitem[Zhao et~al.(2023)Zhao, Li, Liu, Zhang, Sun, and Xu]{zhao2023survival}
Ziqi Zhao, Wentao Li, Ping Liu, Aili Zhang, Jianqi Sun, and Lisa~X Xu.
\newblock Survival analysis for multimode ablation using self-adapted deep
  learning network based on multisource features.
\newblock \emph{IEEE Journal of Biomedical and Health Informatics}, 28\penalty0
  (1):\penalty0 19--30, 2023.

\bibitem[Zhou and Chen(2023)]{zhou2023cross}
Fengtao Zhou and Hao Chen.
\newblock Cross-modal translation and alignment for survival analysis.
\newblock In \emph{Proceedings of the IEEE/CVF International Conference on
  Computer Vision}, pages 21485--21494, 2023.

\bibitem[Zhouyou et~al.(2024)Zhouyou, Lifan, Jiaqi, Deng, Yuanjun, Hao, and
  Yan]{zhouyou2024cmib}
Lai Zhouyou, Long Lifan, Cui Jiaqi, Xiong Deng, Liu Yuanjun, Yin Hao, and Wang
  Yan.
\newblock Cmib: A cross modal interaction with information bottleneck framework
  for multimodal survival analysis.
\newblock In \emph{2024 IEEE International Conference on Cybernetics and
  Intelligent Systems (CIS) and IEEE International Conference on Robotics,
  Automation and Mechatronics (RAM)}, pages 526--530. IEEE, 2024.

\bibitem[Zhu et~al.(2021)Zhu, Yang, Xu, Wang, Zhang, and Han]{zhu2021transfer}
Qi Zhu, Carl Yang, Yidan Xu, Haonan Wang, Chao Zhang, and Jiawei Han.
\newblock Transfer learning of graph neural networks with ego-graph information
  maximization.
\newblock \emph{Advances in Neural Information Processing Systems},
  34:\penalty0 1766--1779, 2021.

\end{thebibliography}
}


\end{document}